\pdfoutput=1

\documentclass[11pt]{article}

\usepackage[final]{acl}

\usepackage{times}
\usepackage{latexsym}

\usepackage[T1]{fontenc}

\usepackage[utf8]{inputenc}

\usepackage{microtype}

\usepackage{inconsolata}

\usepackage{graphicx}
\usepackage{booktabs}
\usepackage{natbib}
\usepackage{multirow}
\usepackage{subcaption}
\usepackage{hyperref}
\usepackage{amsmath}
\usepackage[skins,xparse,breakable,most]{tcolorbox}
\usepackage{fvextra}
\usepackage{paralist}

\usepackage{amssymb}
\usepackage{pifont}
\newcommand{\cmark}{\textcolor{green}{\ding{51}}}%
\newcommand{\xmark}{\textcolor{red}{\ding{55}}}

\newcommand{\dataset}{{MediTOD}}

\usepackage{todonotes}

\title{\dataset{}: An English Dialogue Dataset for Medical History Taking with Comprehensive Annotations}

\author{
 \textbf{Vishal Vivek Saley\textsuperscript{1}},
 \textbf{Goonjan Saha\thanks{Work done when authors were at IIT Delhi.}\textsuperscript{1}},
 \textbf{Rocktim Jyoti Das\footnotemark[1]\textsuperscript{3}},
\\
 \textbf{Dinesh Raghu\textsuperscript{2}},
 \textbf{Mausam\textsuperscript{1}}
\\
 \textsuperscript{1}Indian Institute of Technology, Delhi\\
 \textsuperscript{2}IBM Research, New Delhi, India\\
 \textsuperscript{3}MBZUAI
\\
Vishal.Vivek.Saley@cse.iitd.ac.in, saha.goonjan@gmail.com \\
        rocktimjyotidas@gmail.com, diraghu1@in.ibm.com, mausam@cse.iitd.ac.in
}

\begin{document}
\maketitle
\begin{abstract}


Medical task-oriented dialogue systems can assist doctors by collecting patient medical history, aiding in diagnosis, or guiding treatment selection, thereby reducing doctor burnout and expanding access to medical services. However, doctor-patient dialogue datasets are not readily available, primarily due to privacy regulations. Moreover, existing datasets lack comprehensive annotations involving medical slots and their different attributes, such as symptoms and their onset, progression, and severity. These comprehensive annotations are crucial for accurate diagnosis. Finally, most existing datasets are non-English, limiting their utility for the larger research community.

In response, we introduce \dataset{}, a new dataset of doctor-patient dialogues in English for the medical history-taking task. Collaborating with doctors, we devise a questionnaire-based labeling scheme tailored to the medical domain. Then, medical professionals create the dataset with high-quality comprehensive annotations, capturing medical slots and their attributes. We establish benchmarks in supervised and few-shot settings on \dataset{} for natural language understanding, policy learning, and natural language generation subtasks, evaluating models from both TOD and biomedical domains. We release \dataset{} resources for future research.
\end{abstract}

\section{Introduction}

Medical task-oriented dialogue (TOD) systems are gaining importance in modern healthcare by assisting doctors in patient history-taking, diagnosis suggestions, and treatment recommendations, alleviating doctor burnout and extending the reach of medical services \citep{valizadeh2022ai,kearns2019systematic,laranjo2018conversational}. Recently, medical TOD systems have witnessed significant progress, particularly in individual sub-modules such as natural language understanding (NLU) \citep{Zhang2020MIEAM}, policy learning (POL) \citep{Tchango2022DDXPlusAN}, and natural language generation (NLG) \citep{Yan2021ReMeDiRF}. 
Most of the existing medical TOD datasets contain annotations required for training only one sub-component \citep{Wei2018TaskorientedDS, He2020MedDialogTL, fansi2022towards}, and only a few include annotations for all sub-components, thereby enabling the construction of a complete dialogue system \citep{Yan2021ReMeDiRF, Chen2022ABF}.

\begin{figure}[t]
    \centering
    \includegraphics[width=0.48\textwidth]{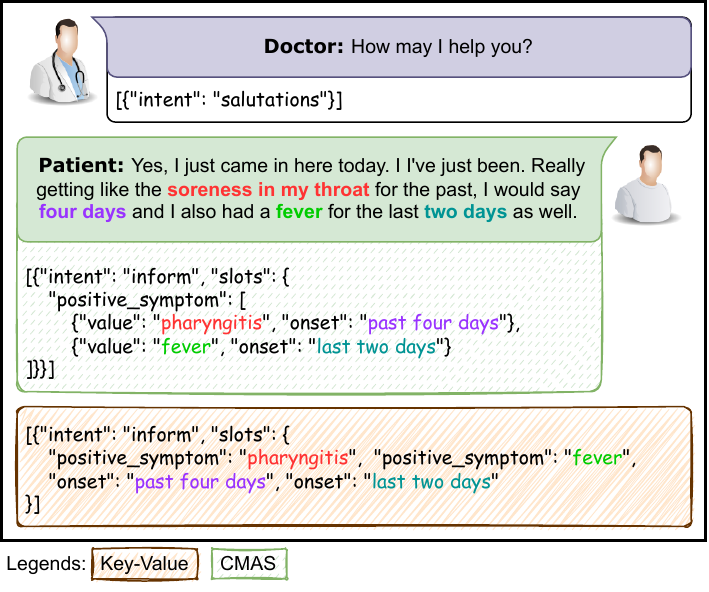}
    \caption{A dialogue turn with annotations labeled using as Comprehensive Medical Attribute Schema (CMAS) in \dataset{}, compared to key-value pairs.}\label{fig:intro_dialog} 
\end{figure}

Training NLU, POL, and NLG sub-modules requires dialogues to be annotated with intents, slots, dialogue states, and actions. In existing medical TOD datasets, slots are primarily represented as key-value pairs. However, this simplistic representation often fails to capture the inherent complexity of the medical domain. For example, in Figure \ref{fig:intro_dialog}, the patient expresses two symptoms (\textit{pharyngitis} and \textit{fever}) along with its onset (\textit{past four days} and \textit{last two days}). Existing annotation schemes would fail to capture the symptom-onset link and only represent four independent key-value pairs.

To overcome this problem, we define a new slot schema, named Comprehensive Medical Attribute Schema (CMAS), that captures the inherent complexity of the slots in the medical domain. It maintains multiple attributes specific to each slot type for better representation. For example in Figure \ref{fig:intro_dialog}, `onset' is treated as an attribute of the slot `symptom',  establishing a more accurate patient profile.

This paper presents \dataset{}, the first publicly available \textit{English} medical TOD dataset annotated in the CMAS ontology.
\dataset{} comprises dialogues from staged doctor-patient interviews in objective structured clinical examination format \citep{harden1979assessment, fareez2022dataset}. By leveraging these high-quality dialogues, privacy concerns are mitigated while providing realistic medical scenarios.
Collaborating closely with doctors, we develop a questionnaire-based annotation framework to collect slots and corresponding attributes.
The annotations are further canonicalized, where possible, to precise medical concepts in Unified Medical Language System (UMLS). Through the release of \dataset{}, we aim to provide a valuable resource for advancing research in medical TOD systems. Our main contributions are as follows.
\begin{compactenum}
    \item We release \dataset{}, a dataset of doctor-patient dialogues with 22,503 utterances annotated, in collaboration with doctors, using a questionnaire-based labeling scheme designed for the medical domain. 
    \item To label utterances in \dataset{}, we develop an annotation portal based on our questionnaire-based scheme. We release this portal alongside the dataset and invite researchers to contribute further to the dataset, enhancing its richness and diversity.
    \item We establish baselines in supervised and few-shot settings for NLU, POL, and NLG TOD tasks on \dataset{} dataset by evaluating representative models from TOD and bio-medical literature. Our results showcase the challenging nature of the dataset.
\end{compactenum}
We make \dataset{} resources publicly available at \url{https://github.com/dair-iitd/MediTOD}.


\section{Related Work}
\begin{table*}[]
\centering
\small
\resizebox{\textwidth}{!}{%
\begin{tabular}{l|c|ccc|cc}
\toprule
\multirow{2}{*}{\textbf{Datasets}} & \multirow{2}{*}{\textbf{Language}} & \multicolumn{3}{c|}{\textbf{Annotations}} & \textbf{\#utterances/} & \multirow{2}{*}{\textbf{\#utterances}} \\ \cmidrule{3-5}
 &  & \textbf{All TOD Tasks} & \textbf{Comprehensive} & \textbf{Canonicalized} & \textbf{dialogue} &  \\ \midrule
CMDD \citep{Lin2019EnhancingDS} & zh & \xmark & \xmark & \cmark & 42.09 & 87,000 \\
MSL \citep{Shi2020UnderstandingMC} & zh & \xmark & \xmark & \cmark & NA & 2,652 \\
MIE \citep{Zhang2020MIEAM} & zh & \xmark & \xmark & \cmark & 16.26 & 18,212 \\
IntRec   \citep{rojowiec2020intent} & de & \xmark & \xmark & \cmark & 57.071 & 2,397 \\
ReMeDi \citep{Yan2021ReMeDiRF} & zh & \cmark & \xmark & \xmark & 16.34 & 25,446 \\
DialoAMC \citep{Chen2022ABF} & zh & \cmark & \xmark & \cmark & 40.02 & 1,64,731 \\
Code-Mixed \citep{dowlagar2023code} & te,en & \cmark & \xmark & \xmark & 9.75 & 29,294 \\ \midrule
\dataset{} (Ours) & en & \cmark & \cmark & \cmark & 95.57 & 22,503 \\ \bottomrule
\end{tabular}%
}
\caption{Publicly available medical dialogue datasets with annotations. \dataset{} is the only English dataset that features both comprehensive (capturing slots and their low-level attributes together) and canonicalized annotations. The language codes "en," "zh," "de," and "te" represent English, Chinese, German, and Telugu, respectively.}
\label{tab:datasets}
\end{table*}

\paragraph{Task-Oriented Dialogue (TOD) Systems: }
General domain TOD systems that assist users in completing tasks such as restaurant table reservation and flight booking often follow a modular design, consisting of three modules -- natural language understanding (NLU), dialogue policy learning (POL), and natural language generation (NLG) \citep{young2013pomdp, wen2016network}. In recent years, there has been significant progress in the field, majorly due to the availability of publicly accessible datasets with dialogue acts annotations \citep{budzianowski2018multiwoz, rastogi2020towards, byrne2019taskmaster, asri2017frames}. 

With pre-trained language models (LMs), recent approaches showcase remarkable performance on all three TOD sub-tasks. \citet{lee2021dialogue}, \citet{zhao2022description}, and \citet{bang2023task} achieve state-of-the-art performance for understanding user's requirements (NLU). \citet{wu2023diacttod}, \citet{bang2023task}, and \citet{sun2022mars} showcase similar trends for system action prediction and response generation tasks. In line with this trend, we benchmark pre-trained language models Flan-T5 \citep{chung2024scaling}, BioGPT \citep{luo2022biogpt} and PPTOD \citep{su2021multi} on \dataset{} dataset.



\paragraph{Medical Dialogue Systems: }
Many such datasets exist; however, only a few (see Table \ref{tab:datasets}) have been annotated. Early works focus on NLU and extract symptom slot-values and their status from a doctor-patient dialogue. CMDD \citep{Lin2019EnhancingDS} and SAT \citep{Du2019ExtractingSA} datasets study this task as sequence labeling, where dialogues are collected from online healthcare forum and clinical setting, respectively. Subsequent datasets, MIE \citep{Zhang2020MIEAM}, ReMeDi \citep{Yan2021ReMeDiRF}, DialoAMC \citep{Chen2022ABF} and Code-Mixed \cite{dowlagar2023code}, introduce additional slots, such as medical test and surgery and collect novel data for the task. 

Notably, ReMeDi and Code-Mixed are the only datasets that collect low-level attributes in their labels. However, \dataset{} differs from them in several ways. First, while these datasets capture the attributes, they do not link them to appropriate slots. For example, ReMeDi would label the onset in a patient's utterance as (time, onset, past two days) without linking it to the symptom, fever. In contrast, \dataset{} uses CMAS to record slots and attributes together through a questionnaire-based annotation framework. Second, \dataset{} has canonicalized values for medical labels, such as symptoms and diseases, to ensure meaningful evaluation and to support future research.

Unlike medical dialogue systems, summarization involves converting doctor-patient dialogues into notes/reports \citep{joshi2020dr, krishna2021generating}. Many synthetic \citep{chintagunta2021medically, wang2023notechat} and human \citep{abacha2023empirical} datasets exist for this task. While summarization is not within the scope of this work, \dataset{} has the potential for adaptation to this task in the future. 

\section{The \dataset{} Dataset}\label{sec:dataset_creation}


To advance research in medical dialogue systems, datasets which capture \textit{canonicalized}, \textit{comprehensive} annotations must be available \textit{publicly}. However, existing datasets (listed in Table \ref{tab:datasets}) only fulfill a subset of these requirements. Moreover, these datasets are often limited to a single demographic, which restricts their broad applicability.


To address these gaps, we curate \dataset{}, an English dataset of doctor-patient dialogues for collecting patient medical histories. First, we form the dialogues in \dataset{} by collecting publicly available transcripts of doctor-patient encounters \citep{fareez2022dataset}. To capture the complexity of the slots, we define a comprehensive medical attribute schema (CMAS) and  develop a questionnaire-based labeling framework to annotate dialogues based on CMAS. For each slot type (e.g., symptom, personal medical history), medical professionals vet questions to capture associated attributes (severity, onset, etc.). Utilizing this framework, professional annotators then label utterances by answering the questionnaire corresponding to each slot under the doctor's supervision. Finally, we canonicalize the slots values to standard medical concepts in the UMLS vocabulary.\footnote{\tt https://uts.nlm.nih.gov/uts/umls/home}

\subsection{Dialogue Acquisition}
Recently, \citet{fareez2022dataset} released a dataset of doctor-patient interviews from five specialties. These interviews involve staged interactions where medical professionals assume the roles of doctor and patient. The doctor systematically gathers healthcare information from the patient following the Objective Structured Clinical Examinations (OSCE) format, covering aspects such as a history of present illness, past medical records, and family health history crucial for diagnosis. 

Notably, \citet{fareez2022dataset} make their dataset publicly available for academic use. Further, their dialogues are highly conversational, averaging 95 utterances per dialogue. Unfortunately, the dataset is not annotated with the necessary labels for building a medical TOD system. In response, we form \dataset{} by labeling 22,503 utterances from the respiratory and musculoskeletal specialties available in their dataset.

\subsection{The CMAS Format}
A doctor-patient dialogue consists of complex slot types such as symptoms, patient medical history, and patient’s habits. 
Symptoms have attributes such as onset (see Figure \ref{fig:intro_dialog}), duration, location, frequency, severity, and progression. Capturing the relationship between these attributes is crucial for creating accurate patient profiles and reliable diagnoses. As explained earlier, existing TOD datasets use key-value pairs to represent both slots and attributes, thus missing the links between them.

\begin{figure*}
    \centering
    \includegraphics[width=0.90\textwidth]{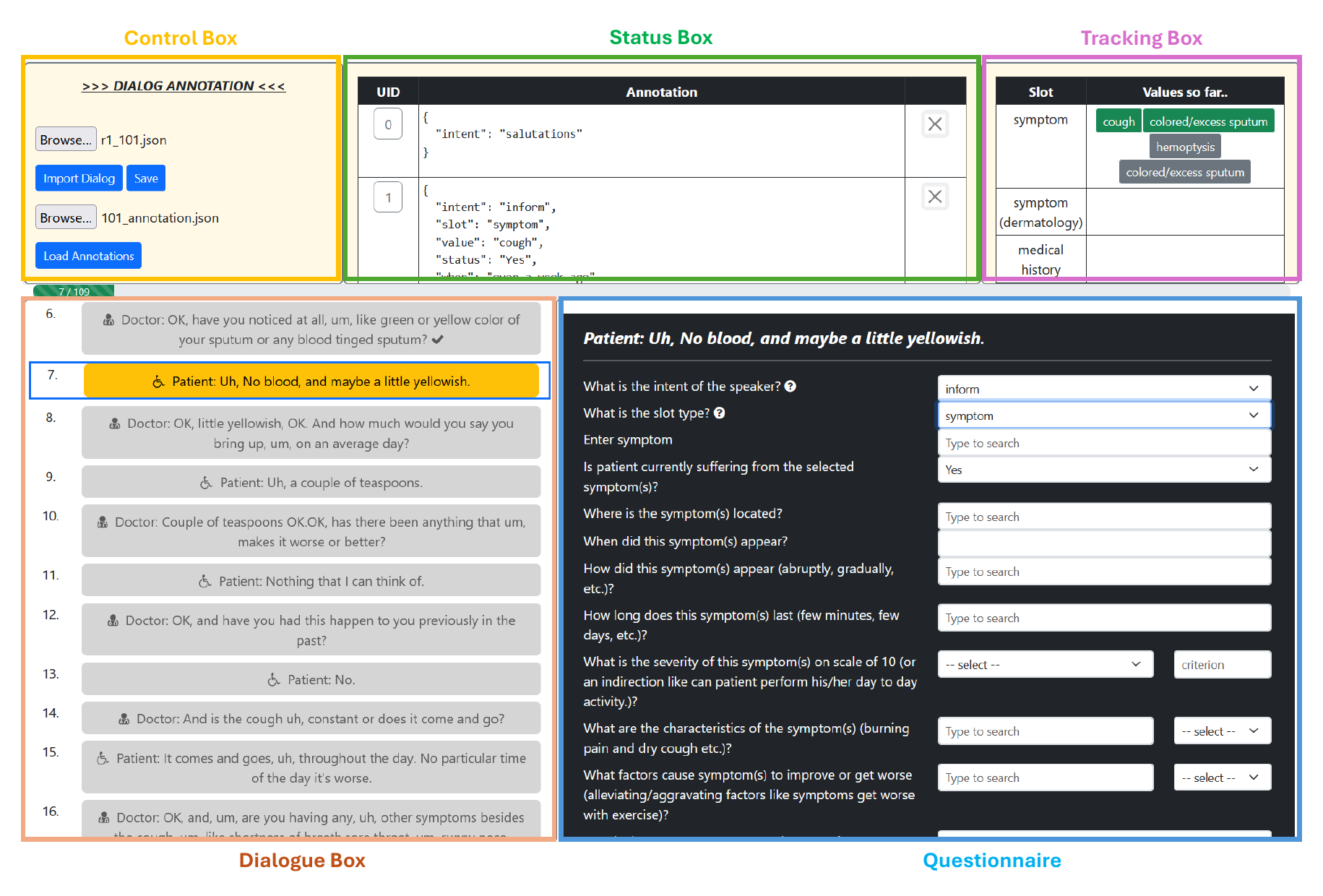}
    \caption{Labeling Interface for questionnaire-based labeling scheme.}\label{fig:ann_portal}
    \label{fig:low_data}
\end{figure*}

In response, for \dataset{}, we develop a Comprehensive Medical Attribute Schema (CMAS) to capture the inherent nature of slots and a questionnaire-based annotation framework that effectively captures the relationship between slots and their attributes while simplifying the labeling task. We ask doctors to design questions for each attribute in a slot. For example, questions like \emph{``Where is the symptom located?"} and \emph{``When did the symptom appear?"} are suggested for the location and onset of symptoms. We provide further details on the questionnaire in Appendix \ref{app:slotwise_questions}.

Using these questions, we develop detailed annotation guidelines for our labeling task that aims to capture intent, slots, and their attributes for each utterance in the dialogues, considering both the utterance itself and the dialogue history. In the next section, we discuss our labeling interface that naturally collects associated attributes.

Table \ref{tab:ontology} lists intents and slot types in CMAS. Table \ref{tab:slot_and_values} in Appendix \ref{app:slotwise_questions} reports the attributes associated with each slot type in the schema.

\begin{table}[]
\centering
\resizebox{0.48\textwidth}{!}{%
\begin{tabular}{l|l}
\toprule
Intents                & Inquire, Inform, Diagnose, Salutations, Chit-chat, Other     \\ \midrule
\multirow{3}{*}{Slots} & Symptom, Patient Medical History, Family Medical History     \\
                       & Habits, Exposure, Medication, Medical Test, Disease          \\
                       & Travel Information, Occupation, Residence, Basic Information \\ \bottomrule
\end{tabular}%
}
\caption{\dataset{} intents and slots}
\label{tab:ontology}
\end{table}

\subsection{Labeling Interface}
Our labeling interface displays a doctor-patient dialogue for annotation. For each utterance, annotators select one or more appropriate intents and slot types. Then, the questionnaires for each chosen slot appear for the annotators to answer. On submission, a status box displays the answers for reference. 
We further enhance usability with editing features, keyboard shortcuts, and a tracking box highlighting the patient's current slot values. Figure \ref{fig:ann_portal} shows a snapshot of our labeling interface. We include the user guide with our annotation guidelines in Appendix \ref{app:annotation_guidelines}. 

The labeling interface offers two significant advantages. First, it requires annotators to capture attributes along with each slot annotated. Specifically, annotators need to provide the slot and its value before recording any additional attributes, which ensures attributes are always linked to their slots and do not exist independently. Second, the interface displays the complete questionnaire for a selected slot, reducing the need for annotators to memorize the slot-attribute relationships.

\subsection{Labeling Process and Quality Control}

We avail a professional annotation service to hire six annotators with medical sciences or pharmacy backgrounds to label the dialogues under the supervision of a doctor.\footnote{The doctor has a professional medical degree with two and half years of hospital experience.} We train annotators by providing them with our detailed annotation guidelines and an example of an annotated dialogue for their reference. Once familiarized with the task and the labeling interface, we ask the annotators to label a different sample dialogue independently. Based on their responses, we offer feedback and point out any issues that need to be addressed. To ensure their understanding, we present a third sample dialogue for them to label independently. During their training, annotators are tested twice, covering 200+ utterances across two dialogues.

We then divide all available dialogues into six groups. Each trained annotator independently labels the assigned group.  To ensure label quality, we systematically introduce seed dialogues into all the groups so that each seed dialogue gets labeled by a pair of annotators. We use the seed dialogue labels to periodically measure inter-annotator agreement and identify any quality issues. Throughout the process, annotators can raise concerns in real time, resolving any ambiguities. Our labeling process has a strong inter-annotator agreement\footnote{We measure inter-annotator agreement between pairs of annotators and report the average.} with Cohen-Kappa $\kappa=0.94$ for intents and $\kappa=0.72$ for slot-value pairs, indicating strong label consistency.

\subsection{Post-Processing}
In post-processing, we examine our collected labels and identify any utterances that may lack complete labels. For instance, in a few utterances such as \textit{I used over-the-counter medicine at night time to help sleep}, the labels did not indicate the \textit{status} of whether the patient had taken the medication. For such cases, which amount to 0.7\% (165 samples) of all the utterances, we ask annotators to review and rectify the labeling to ensure accuracy and completeness. 

Medical concepts, such as symptoms and diseases, often exhibit different surface (and layman) forms. For instance, phrases \textit{shortness of breath} and \textit{difficulty breathing} both refer to the medical concept \textit{Dyspnea}. To ensure consistent medical terminology across annotators, we canonicalize the medical terms in our labels. 
Specifically, we link medical terms to their precise medical concept in UMLS Metathesaurus. UMLS has a large-scale collection of medical vocabularies that facilitate a standardized framework for representing and linking biomedical concepts. Linking to UMLS allows meaningful evaluations and also paves the way for dialogue systems grounded in large-scale online medical databases.

We first divide the slots (and attributes) into two categories - medical and non-medical. Medical ones include symptoms, patient medical history, family medical history, etc. Non-medical ones include duration, frequency, residence, etc. 

To canonicalize medical slots/attributes, we use QuickUMLS\footnote{https://github.com/Georgetown-IR-Lab/QuickUMLS} string matching to generate a set of candidate UMLS concepts. Then, we manually verify these candidates to filter extraneous ones, considering surface forms and the context within the dialogue. We keep the doctor in the loop to review the candidates and pick the final concept, provide corrections or recommendations where necessary, and resolve any ambiguities. Finally, we replace the surface forms with their corresponding canonicalized version to ensure that the medical terminology aligns with professional standards.

In contrast to medical slots/attributes, non-medical ones lack standardized vocabularies of concepts and are thus not canonicalized in our dataset.





\subsection{Dataset Statistics}
\begin{table}[]
\centering
\small
\resizebox{0.5\textwidth}{!}{%
\begin{tabular}{ccccc}
\toprule
\textbf{Split} & \textbf{\#dialogues} & \textbf{\#utterances} & \textbf{Avg \#utterances} & \textbf{Avg \#words} \\
 &  &  & per dialogue & per utterance \\ \midrule
Train & 175         & 16,852        & 95.29                         & 13.46                     \\
Valid & 20          & 1,869         & 92.45                         & 12.77                     \\
Test  & 18          & 1,798         & 98.89                         & 13.03                     \\
Out-of-domain Test  & 20        & 2,197         & 109.85                        & 14.00                     \\ \midrule
Total & 213         & 22,503        & 96.57                         & 13.42                     \\ \bottomrule
\end{tabular}%
}
\caption{\dataset{} statistics.}\label{tab:data_stats}
\end{table}

We treat respiratory and musculoskeletal dialogues in \dataset{} separately. We use dialogues from the respiratory specialty for model building and in-domain benchmarking, while dialogues from the musculoskeletal specialty serve as an out-of-domain test set.

We divide the respiratory dialogues into train, validation, and test sets. First, we form a high-quality in-domain test set consisting of the seed dialogues we used for quality control. Each test dialogue is thus labeled by two different annotators. To obtain the final labels, we task a third annotator to resolve the inconsistencies between the two sets of labels. Finally, we randomly split the remaining respiratory dialogues into train and validation sets.

The out-of-domain test set consists of musculoskeletal dialogues. A single annotator labeled all these dialogues. To ensure quality, we conducted periodic checks on the submitted labels and provided feedback to the annotator as needed. Table \ref{tab:data_stats} lists overall statistics for \dataset{} dataset. With 22,503 annotated utterances, \dataset{} enables meaningful training and evaluation of the machine learning models. Dialogues in \dataset{} are highly conversational, with an average of 96 utterances per dialogue. 

\begin{figure}[]
    \centering
    \begin{subfigure}[t]{0.23\textwidth}
    \includegraphics[height=1.6in]{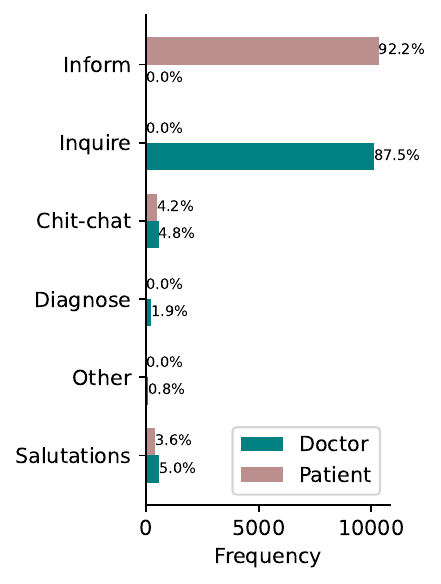}
    \end{subfigure}
    \begin{subfigure}[t]{0.23\textwidth}
    \includegraphics[height=1.6in]{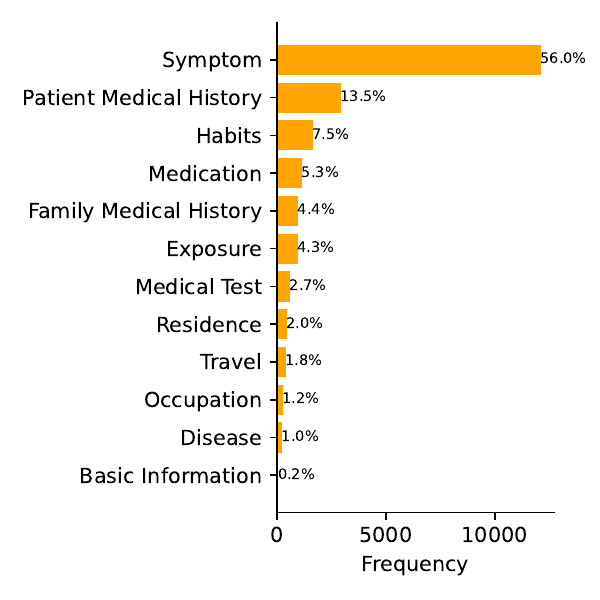}
    \end{subfigure}
    \caption{Distribution of utterances \dataset{} dataset over intents (left) and slots (right).}\label{fig:int_slot_dist}
\end{figure}

Figure \ref{fig:int_slot_dist} shows the distribution of utterances in \dataset{} over different intents and slot values. For doctors, \textit{Inquire} accounts for 87.5\% of the total utterances, followed by \textit{Salutations} (5.0\%), \textit{Chit-chat} (4.8\%), \textit{Diagnose} (1.9\%), and \textit{Other} (0.8\%). For patients, \textit{Inform} (92.2\%) is the dominant slot, followed by \textit{Chit-chat} (4.2\%) and \textit{Salutations} (3.6\%). Similarly, 56\% of the utterances discuss \textit{Symptom} followed by \textit{Patient Medical History} (13.5\%) and \textit{Habits} (7.5\%).

\begin{figure}[]
    \centering
    \includegraphics[width=0.48\textwidth]{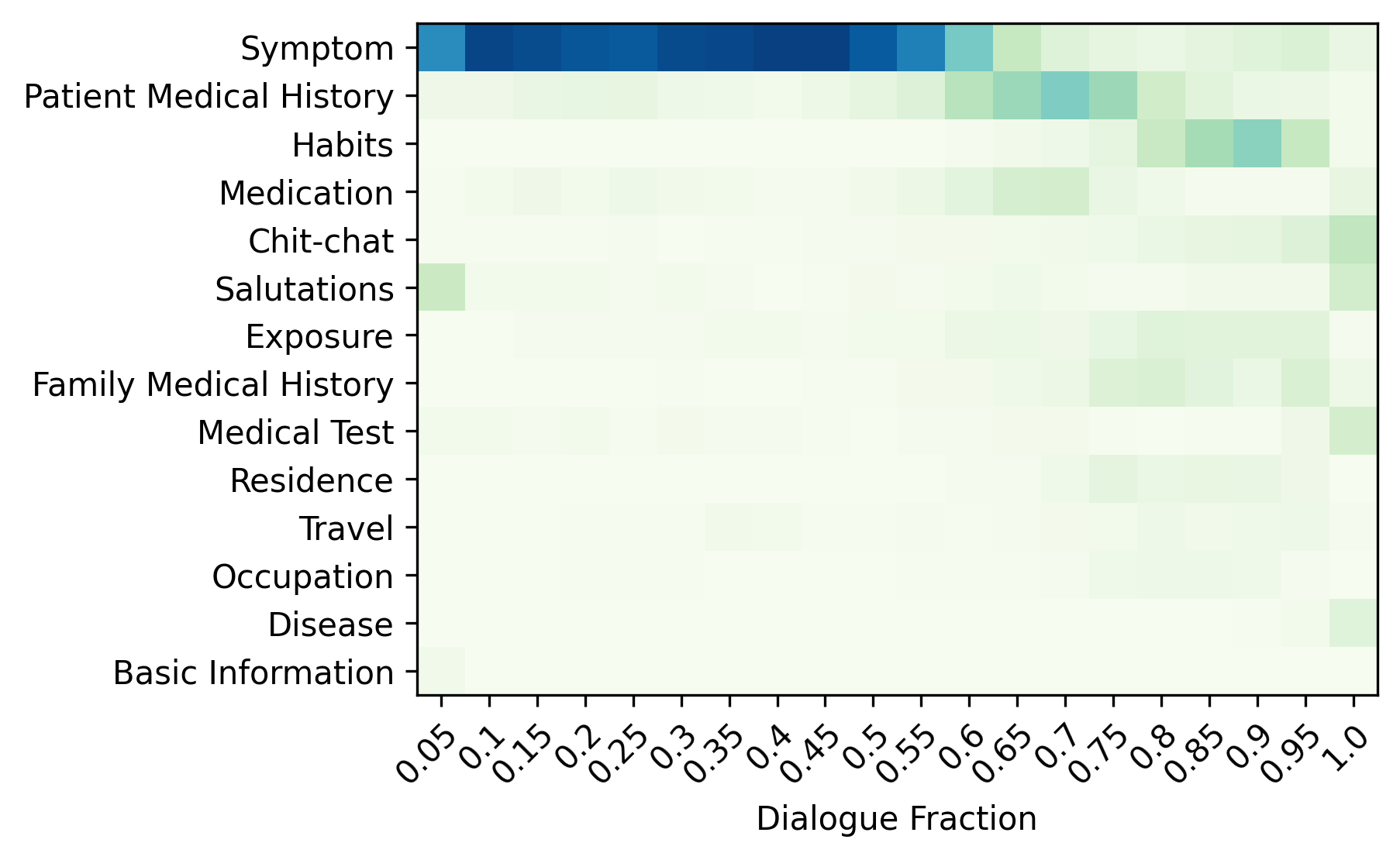}
    \caption{dialogues in \dataset{} dataset are systematic.}\label{fig:dialog_struc}
\end{figure}

Dialogues in \dataset{} are highly systematic. To elucidate that, we divide each dialogue into ten equal segments. For each segment, we find the distribution of the slots across all dialogues. Figure \ref{fig:dialog_struc} shows the resultant heatmap. During the first half of their conversation, the doctor and the patient primarily discuss symptoms. Then, they transition into other slots, such as patient medical history, habits, medication, etc.

\section{Experimental Setup}


\dataset{} supports all three subtasks in a TOD system -- natural language understanding (NLU), policy learning (POL), and natural language generation (NLG). Table \ref{tab:tasks} (appendix) illustrates representative examples from each subtask.

NLU involves understanding the information presented in the latest patient utterance. Specifically, given the dialogue so far, an NLU model predicts the intent, active slots, and their attributes of the patient utterance. This updates the dialogue state, i.e., the aggregate information the patient reports up to the current turn in the dialogue.


POL requires predicting the doctor's next action. 
An action consists of intents, slots, and associated attributes. A POL model inputs the dialogue history and the dialogue state and predicts the action.
Finally, NLG involves transforming the doctor's action into natural language. An NLG model predicts the doctor's utterance based on the dialogue history and the action from the POL model.


\subsection{Evaluation Metrics}
For NLU, we compute an F1 score by matching intent, slots, and associated attributes from gold and predicted labels. Before matching, we unroll the gold and predicted NLU labels into sets of the form \emph{\{(intent, slot, value, attribute, attribute-value)\}}. For instance, the NLU label in Figure \ref{fig:intro_dialog} is transformed to \emph{{[}(inform, positive symptom, pharyngitis), (inform, positive symptom, pharyngitis, onset, past four days), (inform, positive symptom, fever), (inform, positive symptom, fever, onset, last two days){]}}. 



In \dataset{}, medical attributes are canonicalized, so we simply use exact match scores for them. However, for non-medical attributes, where different strings can have the same meaning, just string matching is too conservative. E.g., `2 days ago' and `two days before' convey the same meaning, but will get counted as non-matches. For non-medical attributes, we use ChatGPT \citep{openai2024gpt35turbo} to adjudicate semantic equivalence and use that for F1 score computation. Our prompt is in Appendix \ref{app:metric_prompt}.  In  experiments, we report medical and non-medical scores separately, in addition to overall scores.


For POL, we could use precision, recall, and F1 scores, similar to NLU; however, in conversation, it is quite possible for the doctor to change the order of questions somewhat. 
To account for this, we use the Precision@K metric. At a given test turn, we check if the medical attributes predicted from the POL model are present in the gold actions within the next K turns. We report Precision@K for K = 1, 4, 8, and infinity. Finally, for NLG, we use BLEU \citep{papineni2002bleu}, Rouge \citep{lin-2004-rouge}, and BERTScore \citep{zhang2019bertscore} to measure the generation quality.



\subsection{Baselines}
We model all three tasks as \emph{seq2seq} learning and evaluate baselines in the supervised and in-context learning settings. Baselines use a respiratory (in-domain) dataset for development and are evaluated using both an in-domain (respiratory) test set and an out-of-domain (musculoskeletal) test set.

\vspace{1ex}
\noindent \textbf{Supervised.} We fine-tune several pre-trained language models on the \dataset{} training set for the three tasks. We utilize the PPTOD (base) \citep{su2021multi} and Flan-T5 (base) \citep{chung2024scaling} encoder-decoder transformer models, pre-trained on general domain TOD tasks and the Flan suite of tasks, respectively. Additionally, we employ BioGPT (medium) \citep{luo2022biogpt}, a decoder-only transformer model pre-trained on extensive biomedical literature. Finally, we fine-tune Llama3 8B Instruct \citep{llama3modelcard} (referred to as Llama3 henceforth) and OpenBioLLM 8B \citep{OpenBioLLMs} models, which serve as representatives of large language models (LLMs). After fine-tuning, we evaluate the models on the \dataset{} test set.

\vspace{1ex}
\noindent \textbf{In-context Learning.} We prompt several large language models (LLMs) to make predictions on test samples across the three tasks. For each test sample, we select the top five exemplars \footnote{We limit the selection to five exemplars to ensure the prompt stays within the 4096 token limit.} from the training set whose dialogue histories are semantically closest to the test sample \citep{liu-etal-2022-makes}. To identify the top exemplars, we use the BAAI/bge-large-en-v1.5 model \citep{bge_embedding} to encode the dialogue history and perform a maximum inner product search. Sample prompts used in our experiments are provided in Appendix \ref{app:prompts}. We evaluate the performance of Llama3 8B/70B Instruct \citep{llama3modelcard}, OpenBioLLM 8B/70B \citep{OpenBioLLMs}, ChatGPT (gpt-3.5-turbo-0125) \citep{openai2024gpt35turbo}, and GPT-4 (gpt-4-1106-preview) and GPT-4-Turbo (gpt-4-turbo-2024-04-09) \citep{openai2023gpt4preview}.

\subsection{Implementation Details}
We adapted a publicly available codebase for the PPTOD model to \dataset{}\footnote{\href{https://github.com/awslabs/pptod}{https://github.com/awslabs/pptod}}. For training this model, we utilized a learning rate of $1e-3$ and a batch size of 64. For the Flan-T5 and BioGPT models, we employed a learning rate of $1e-4$ and a batch size of 16. These models were trained on a single V100 GPU with 32 GB of memory.

We train Llama3 with Unsloth\footnote{\url{https://github.com/unslothai/unsloth}} to enhance the efficiency of both training and inference. We fine-tuned the Llama3 using LoRA \citep{hu2021lora} and use parameters $r=32$ and $\alpha=32$, a learning rate of $1e-4$, a batch size of 16, and a cosine learning rate scheduler with a warm-up period of 10\% for both models. This training was performed on two A100 40GB GPUs, taking approximately five hours to complete. We trained OpenBioLLM models under similar settings. During inference, we employ greedy decoding (temp. $=$ 0) for all in-context and supervised baselines. 

\section{Results}
We begin by evaluating the models for the NLU, POL, and NLG tasks on the in-domain test set. Next, we assess the top-performing models from the in-domain evaluation on the out-of-domain dataset.

\subsection{Natural Language Understanding}

\begin{table}[]
\centering
\resizebox{0.495\textwidth}{!}{%
\begin{tabular}{llccc}
\toprule
& & \textbf{Overall} & \textbf{Medical} & \textbf{Non-Medical}\\
& \multirow{-2}{*}{\textbf{Model}} & \textbf{F1} & \textbf{F1} & \textbf{F1} \\ \midrule
\multirow{5}{1em}{\rotatebox[origin=c]{90}{Supervised}} & PPTOD (base) & 0.6849 & 0.7268 & 0.5141 \\
& Flan-T5 (base) & 0.6887 & 0.7354 & 0.5062 \\
& BioGPT & 0.6090 & 0.6612 & 0.4187 \\
& OpenBioLLM 8B & 0.6731 & 0.7294 & 0.4791 \\
& Llama3 8B & \textbf{0.7139} & \textbf{0.7603} & \textbf{0.5397} \\
\midrule
\multirow{6}{1em}{\rotatebox[origin=c]{90}{In-context}} & OpenBioLLM 8B & 0.6056 & 0.6431 & 0.4416 \\
& OpenBioLLM 70B & 0.6090 & 0.6499 & 0.4664 \\
& Llama3 8B & 0.5251 & 0.5611 & 0.4048 \\
& Llama3 70B & 0.5736 & 0.6150 & 0.4422 \\
& ChatGPT & 0.5929 & 0.6337 & 0.4425 \\
& GPT-4 & 0.6351 & 0.6715 & 0.5043 \\
& GPT-4-Turbo & \textbf{0.6641} & \textbf{0.6999} & \textbf{0.5329} \\ \bottomrule
\end{tabular}%
}
\caption{In-Domain Model Evaluation for the \dataset{} NLU Task}
\label{tab:nlu_results}
\end{table}

Table \ref{tab:nlu_results} compares the performance of various baselines on an NLU task. At a high level, Llama3 8B achieves the best overall performance among the supervised models. This superior performance can be attributed to its ability to recognize medical and non-medical slots better than its competitors. While most models demonstrate competitive performance in recognizing non-medical attributes, the Llama3 model has a clear advantage for medical attributes, scoring 0.0249 points higher over the nearest Flan-T5 baseline. 

In the in-context learning setting, pre-training on biomedical corpus offers a clear advantage, with OpenBioLLM significantly outperforming Llama3 models from the same weight class. Specifically, OpenBioLLM 8B and OpenBioLLM 70B surpass their Llama3 counterparts by 0.0805 and 0.0354 pts, respectively. OpenBioLLM showcases a superior understanding of medical attributes, compared to the general-purpose Llama models. OpenBioLLM 70B exhibits a slight edge of 0.0174 points over ChatGPT. However, GPT-4-Turbo achieves the best in-context performance overall. While GPT-4-Turbo's non-medical F1 is comparable to supervised models, its performance on medical attributes lags behind the supervised Llama3 model by 0.0573 pts. This highlights the potential for further improvements in in-context learning baselines, particularly in the medical domain.

Tables \ref{tab:nlu_example1} and \ref{tab:nlu_example2} display example responses from different baseline models. In table \ref{tab:nlu_example1}, the models struggle with distinguishing between related but distinct medical concepts, such as confusion and mental fatigue. Extracting multiple slots and attributes also presents a challenge for the models. As shown in table \ref{tab:nlu_example2}, the models either fail to recognize all the symptoms from the input or make errors when linking the attributes. 

\subsection{Policy Learning}

\begin{table}[]
\centering
\small
\resizebox{0.495\textwidth}{!}{%
\begin{tabular}{llccc}
\toprule
& & \textbf{Overall} & \textbf{Medical} & \textbf{Non-Medical}\\
& \multirow{-2}{*}{\textbf{Model}} & \textbf{F1} & \textbf{F1} & \textbf{F1} \\ \midrule
\multirow{5}{1em}{\rotatebox[origin=c]{90}{Supervised}} & PPTOD (base) & 0.2099 & 0.2101 & 0.2069 \\
& Flan-T5 (base) & 0.1999 & 0.2033 & 0.1495 \\
& BioGPT & 0.1870 & 0.1853 & 0.2156 \\
& OpenBioLLM 8B & 0.2085 & 0.2172 & 0.1190 \\
& Llama3 8B & \textbf{0.2389} & \textbf{0.2392} & \textbf{0.2329} \\ \midrule
\multirow{6}{1em}{\rotatebox[origin=c]{90}{In-context}} & OpenBioLLM 8B & 0.1052 & 0.1086 & 0.0336 \\
& OpenBioLLM 70B & 0.1214 & 0.1273 & 0.0247 \\
& Llama3 8B & 0.1072 & 0.1116 & 0.0160 \\
& Llama3 70B & 0.1000 & 0.1042 & 0.0154 \\
& ChatGPT & 0.1099 & 0.1124 & \textbf{0.0536} \\
& GPT-4 & 0.0904 & 0.0937 & 0.0167 \\ 
& GPT-4-Turbo & \textbf{0.1296} & \textbf{0.1346} & 0.0172 \\ \bottomrule
\end{tabular}%
}
\caption{In-domain model performance on the \dataset{} POL task evaluated using the F1 scores.}
\label{tab:pol_results}
\end{table}

\begin{table}[]
\centering
\small
\resizebox{0.495\textwidth}{!}{%
\begin{tabular}{llcccc}
\toprule
& \textbf{Model} & \textbf{P@1} & \textbf{P@4} & \textbf{P@8} & \textbf{P@Inf} \\ \midrule
\multirow{5}{1em}{\rotatebox[origin=c]{90}{Supervised}} & PPTOD (base) & 0.1528 & 0.2839 & 0.3261 & 0.3592 \\
& Flan-T5 (base) & 0.1152 & 0.2041 & 0.2383 & 0.2645 \\
& BioGPT & 0.1345 & 0.2509 & 0.2999 & 0.3478 \\
& OpenBioLLM 8B & 0.1608 & 0.2953 & 0.3375 & 0.3740 \\
& Llama3 8B & \textbf{0.1881} & \textbf{0.3181} & \textbf{0.3820} & \textbf{0.4162} \\ \midrule
\multirow{6}{1em}{\rotatebox[origin=c]{90}{In-context}} & OpenBioLLM 8B & \multicolumn{1}{l}{0.0593} & \multicolumn{1}{l}{0.1220} & \multicolumn{1}{l}{0.1505} & \multicolumn{1}{l}{0.1938} \\
& OpenBioLLM 70B & \textbf{0.0855} & 0.1357 & 0.1619 & 0.1950 \\
& Llama3 8B & 0.0616 & 0.1117 & 0.1357 & 0.1847 \\
& Llama3 70B & 0.0604 & 0.1220 & 0.1505 & 0.1779 \\
& ChatGPT & 0.0832 & \textbf{0.1448} & \textbf{0.1790} & \textbf{0.2201} \\
& GPT-4 & 0.0410 & 0.0798 & 0.1106 & 0.1893 \\
& GPT-4-Turbo & \textbf{0.0855} & 0.1357 & 0.1699 & 0.203 \\ \bottomrule
\end{tabular}%
}
\caption{In-domain model performance on the \dataset{} POL task evaluated using the Precision@K (P@K) scores.}
\label{tab:pol_results2}
\end{table}

Tables \ref{tab:pol_results} and \ref{tab:pol_results2} present the performance of various models on the POL task. Unlike NLU, where performance can vary widely, all models demonstrate competitive results in the supervised setting. The Llama3 8B models are the top performers, with Llama3 gaining 0.0069 points over the other baselines. A similar trend is observed with the Precision@K measure, where the Llama3 model maintains a slight advantage over its competitors across different values of K.

In the in-context learning setting, all baseline models show similar results, with the GPT-4-Turbo model achieving the highest score. However, for the Precision@K measure at K=4, ChatGPT surpasses the GPT-4-Turbo model. Interestingly, ChatGPT performs better than GPT-4 in this task. Upon careful study, we found that GPT-4 generates responses that violate the CMAS label structure in 20.75\% cases. In contrast, ChatGPT and GPT-4-Turbo make such errors in only 3.31\% and 7.07\%.

Notably, the overall performance of in-context baselines is significantly lower than that of supervised models. This disparity arises because policy learning inherently requires models to plan ahead. Supervised models, which learn policy directly from the data, thus have an advantage.

Even though all models behave similarly, the raw scores are not very high, suggesting that more research is needed to improve this component.

\subsection{Natural Language Generation}


\begin{table}[]
\centering
\resizebox{0.495\textwidth}{!}{%
\begin{tabular}{llccccc}
\toprule
& \multirow{2}{*}{\textbf{Model}} & \textbf{BLEU} & \textbf{BLEU} & \textbf{ROUGE} & \textbf{ROUGE} & \textbf{BERT} \\
&  & \textbf{2} & \textbf{4} & \textbf{1} & \textbf{L} & \textbf{Score} \\  \midrule
\multirow{5}{1em}{\rotatebox[origin=c]{90}{Supervised}} & PPTOD (base) & 18.2 & 8.2 & 0.300 & 0.286 & 0.881 \\
& Flan-T5 (base) & 27.2 & 15.3 & \textbf{0.492} & \textbf{0.466} & 0.907 \\
& BioGPT & 24.3 & 13.1 & 0.462 & 0.438 & 0.901 \\
& OpenBioLLM 8B & 28.8 & 16.0 & 0.488 & 0.462 & \textbf{0.908} \\
& Llama3 8B & \textbf{31.4} & \textbf{17.5} & \textbf{0.492} & 0.459 & \textbf{0.908} \\   \midrule
\multirow{6}{1em}{\rotatebox[origin=c]{90}{In-context}} & OpenBioLLM 8B & 19.4 & 8.6 & 0.318 & 0.295 & 0.885 \\
& OpenBioLLM 70B & \textbf{21.5} & \textbf{10.5} & \textbf{0.361} & \textbf{0.325} & 0.889 \\
& Llama3 8B & 18.3 & 7.6 & 0.302 & 0.270 & 0.880 \\
& Llama3 70B & 18.9 & 8.1 & 0.310 & 0.274 & 0.881 \\
& ChatGPT & 20.4 & 8.7 & 0.335 & 0.298 & \textbf{0.890} \\
& GPT-4 & 16.1 & 6.3 & 0.300 & 0.260 & 0.870 \\
& GPT-4-Turbo & 19.3 & 8.1 & 0.327 & 0.289 & 0.889 \\ \bottomrule
\end{tabular}%
}
\caption{In-domain model performance on \dataset{} NLG task.}
\label{tab:nlg_results}
\end{table}


Table \ref{tab:nlg_results} reports NLG results. In the supervised setting, Llama3 8B and OpenBioLLM 8B surpass other baselines, with Llama3 8B achieving the best performance across most of the metrics. In the in-context setting, the OpenBioLLM 70B emerges as a clear winner. Interestingly, few-shot models perform very competitively with supervised models on the BERTScore metric. This suggests that their responses are semantically similar to the gold standard, even if they differ lexically.

We analyze responses from the OpenBioLLM 8B supervised model to identify its shortcomings. The model performs well when asking patients for information, effectively using natural phrases like "Ok, and..." to convey understanding. However, it has difficulty converting multiple actions into natural language, particularly towards the end of conversations. It generates repetitive strings when the doctor discusses possible diagnoses and necessary medical tests or provides support. In contrast, ChatGPT and GPT-4 fare well in such cases.

\subsection{Out-of-domain Evaluation}
We tested the performance of leading supervised models (OpenBioLLM 8B, Llama3 8B) and in-context models (ChatGPT, GPT-4-Turbo) on musculoskeletal dialogues in \dataset{}. The results for NLU, POL, and NLG tasks are shown in Tables \ref{tab:ood_nlu}, \ref{tab:ood_pol}, and \ref{tab:ood_nlg}. For NLU and NLG tasks, the models performed worse on out-of-domain data compared to in-domain data. This is because musculoskeletal dialogues use medical terms that are different from those in the respiratory domain. However, for the POL task, models maintained their in-domain performance, even though it was still low. This suggests that more research is needed to close the performance gap for NLU and NLG tasks and to improve POL task performance overall.

\begin{table}[]
\centering
\resizebox{0.495\textwidth}{!}{%
\begin{tabular}{lcccc}
\toprule
 & \multirow{2}{*}{\textbf{Model}} & \textbf{Overall} & \textbf{Medical} & \textbf{Non-Medical} \\ 
 &  & F1 & F1 & F1 \\ \midrule
\multirow{2}{*}{Supervised} & OpenBioLLM 8B & 0.4747 & 0.5027 & 0.3512 \\
 & Llama 8B & \textbf{0.524} & \textbf{0.5391} & \textbf{0.4431} \\ \midrule
\multirow{2}{*}{In-Context} & ChatGPT & 0.3637 & 0.3953 & 0.2141 \\
 & GPT-4-Turbo & \textbf{0.4481} & \textbf{0.476} & \textbf{0.3176} \\ \bottomrule
\end{tabular}%
}
\caption{Out-of-domain model evaluation on \dataset{} NLU task.}
\label{tab:ood_nlu}
\end{table}

\begin{table}[]
\centering
\resizebox{0.495\textwidth}{!}{%
\begin{tabular}{lcccc}
\toprule
 & \multirow{2}{*}{\textbf{Model}} & \textbf{Overall} & \textbf{Medical} & \textbf{Non-Medical} \\ 
 &  & F1 & F1 & F1 \\ \midrule
\multirow{2}{*}{Supervised} & OpenBioLLM 8B & \textbf{0.2473} & \textbf{0.2518} & 0.1877 \\
 & Llama 8B & 0.2236 & 0.2233 & \textbf{0.2294} \\\midrule
\multirow{2}{*}{In-context} & ChatGPT & 0.0836 & 0.0841 & 0.0702 \\
 & GPT-4-Turbo & \textbf{0.1252} & \textbf{0.1276} & \textbf{0.0722} \\ \bottomrule
\end{tabular}%
}
\caption{Out-of-domain model evaluation on \dataset{} POL task.}
\label{tab:ood_pol}
\end{table}

\begin{table}[]
\centering
\resizebox{0.495\textwidth}{!}{%
\begin{tabular}{lcccccc}
\toprule
 & \multirow{2}{*}{Model} & BLEU & BLEU & ROUGE & ROUGE & BERT \\ 
 &  & 2 & 4 & 1 & L & Score \\ \midrule
\multirow{2}{*}{Supervised} & OpenBioLLM 8B & 25.1 & 13.9 & \textbf{0.486} & \textbf{0.449} & \textbf{0.907} \\
 & Llama 8B & \textbf{26.7} & \textbf{13.9} & 0.469 & 0.429 & 0.904 \\ \midrule
\multirow{2}{*}{In-context} & ChatGPT & 17.8 & \textbf{7.029} & 0.340 & \textbf{0.297} & \textbf{0.893} \\
 & GPT-4-Turbo & \textbf{18.9} & 7.0 & \textbf{0.342} & 0.295 & 0.892 \\ \bottomrule
\end{tabular}%
}
\caption{Out-of-domain model evaluation on \dataset{} NLG task.}
\label{tab:ood_nlg}
\end{table}

\section{Conclusion and Future Works}
In this work, we introduced \dataset{}, a novel English dataset of doctor-patient dialogues for collecting patient medical history. Unlike existing medical datasets, \dataset{} uses a novel schema (CMAS) to capture attributes such as the onset and duration of symptoms relevant for downstream diagnosis. Further, we link values for medical attributes in \dataset{} labels to their precise medical concepts within UMLS vocabularies. Finally, we propose new benchmarks for NLU, POL, and NLG tasks in the medical dialogue domain. Our initial experiments with baseline models reveal the challenges inherent in these tasks and underscore the potential for improvement.

Furthermore, \dataset{} facilitates the exploration of additional research settings such as Knowledge Grounded TOD. Canonicalization in \dataset{} allows for seamless integration of UMLS vocabularies and Semantic networks into TOD settings, potentially enhancing performance. \dataset{} also opens doors to research in medical dialogue summarization, offering opportunities to distill complex dialogues into concise medical summaries. Our annotation portal can aid in creating large-scale medical dialogue datasets for medical specialties beyond pulmonology and musculoskeletal. This expansion will broaden the applicability and relevance of \dataset{} in medical dialogue research. We release the \dataset{} resources at \url{https://github.com/dair-iitd/MediTOD}.

\section*{Ethics Statement}
In this work, we introduce the \dataset{} dataset, which consists of doctor-patient dialogues aimed at gathering patient medical information. This section scrutinizes our data annotation process, as outlined in section \ref{sec:dataset_creation}, from an ethical perspective.

Regarding data sourcing, we thank \citet{fareez2022dataset} for generously providing their data under the Creative Commons CC0 license\footnote{Source dialogues are available \href{https://springernature.figshare.com/articles/dataset/Collection_of_simulated_medical_exams/16550013?backTo=/collections/A_dataset_of_simulated_patient-physician_medical_interviews_with_a_focus_on_respiratory_cases/5545842}{here}.}. The source dialogues in our dataset are simulated interviews conducted by medical professionals, which portray both doctor and patient roles. It's important to note that no actual patient information is disclosed within these dialogues.

We use a professional annotation service specializing in medical data solutions. The provider has been in business for ten years. The labeling process for our dataset involved employing six annotators under the guidance of a doctor. The annotators have backgrounds in medical sciences or pharmacy, ensuring a high level of expertise. The doctor has a professional medical degree and two and half years of hospital experience. Before we started the labeling process, we declared that this work was for scientific advancement and, thus, would be released for public consumption. Each annotator is paid 8 USD per hour, which is above the average wage of data annotators in our country.

We utilize the UMLS Metathesaurus to standardize medical slot values in MediTOD, employing the QuickUMLS software. The National Library of Medicine, Department of Health and Human Services (NLM), grants the UMLS vocabulary license free of charge, which we have obtained for our work. We are committed to properly attributing UMLS and meeting their licensing requirements upon the public release of our dataset.

While releasing MediTOD, we acknowledge its significance in advancing medical dialogue systems. However, it's crucial to emphasize that this data is intended solely for research purposes. We strongly advise against its use in real-life patient consultations or activities that could potentially endanger patients' well-being. Finally, through our work, we want to develop systems for assisting doctors in their work and reducing their burnout. However, we do not claim such a system can work independently without any oversight of healthcare providers.

\section*{Limitations}
While MediTOD makes a meaningful contribution to the medical dialogue community through detailed canonical annotations, it's essential to acknowledge its limitations.
Primarily, MediTOD focuses solely on dialogues from the fields of pulmonology and musculoskeletal. This restricted scope might limit its applicability across other medical specialties. Nonetheless, we're optimistic that our methodology can be adapted to annotate dialogues from different medical fields.
Additionally, the dialogues annotated in MediTOD are exclusively in English. This linguistic limitation may restrict access to the non-English speaking portion of the population. However, we recognize the importance of inclusivity and are open to exploring ways to address language barriers to broaden the reach of our work.

\section*{Acknowledgements}
This work is supported by IBM AI Horizons Network grant, grants by Google, Verisk, and Microsoft, an IBM SUR award and the Jai Gupta chair fellowship by IIT Delhi. Vishal is supported by a Google Fellowship. We thank the IIT Delhi HPC facility for its computational resources. We are grateful to Microsoft AFMR for supporting this work.

\bibliography{custom}

\appendix

\section{\dataset{} Tasks Examples}
\begin{table}[h!]
    \small
    \centering
    \begin{tabular}{p{0.07\textwidth} | p{0.35\textwidth}}
        \toprule
        Dialogue History     &  \parbox{0.35\textwidth}{\fontsize{9pt}{11pt} \selectfont
        \textbf{Doctor:} What brings you in here today?\\ 
        \textbf{Patient:} Um, I'm just, I'm here because I've had this cough for the past two weeks and uh, it's just not going away.\\ 
        \textcolor{blue}{\textbf{Doctor:} Okay, and um, is it getting worse at all really?\\ 
        \textbf{Patient:} Not really, it's just been the same.}}\\ \midrule
        NLU & \fontsize{9pt}{11pt} \selectfont
        \begin{Verbatim}[breaklines]
[{
  "intent": "inform",
  "slots": {
    "positive_symptom": [{
      "value": "coughing",
      "progression": "unchanged with time"
    }]
  }
}]
        \end{Verbatim}
        \\ \midrule
        Patient State & \fontsize{9pt}{11pt} \selectfont
        \begin{Verbatim}[breaklines]
{
  "positive_symptom": [{
    "value": "coughing",
    "onset": "two weeks ago",
    "progression": "unchanged with time"
  }]
}
        \end{Verbatim}
        \\ \midrule
        POL & \fontsize{9pt}{11pt} \selectfont
        \begin{Verbatim}[breaklines]
[{
  "action": "inquire",
  "symptom": [{
    "value": "coughing",
    "checks": [{
      "type": "characteristics",
      "values": ["wet cough or dry cough"]
    }]
  }]
}]
        \end{Verbatim}
        \\ \midrule
        NLG & \fontsize{9pt}{11pt} \selectfont How would you describe the cough? Is it a wet cough or a dry cough? \\ \bottomrule
    \end{tabular}
    \caption{NLU, POL and NLG tasks in \dataset{}. Latest exchange is highlighted in \textcolor{blue}{blue}.}
    \label{tab:tasks}
\end{table}


\section{Additional Results}
Table \ref{tab:nlu_results_ext} and Table \ref{tab:pol_results_ext} present precision, recall and F1 for NLU and POL tasks, respectively.

\begin{table*}[]
\centering
\resizebox{\textwidth}{!}{%
\begin{tabular}{ll|ccc|ccc|ccc}
\toprule
& \multirow{2}{*}{Model} & \multicolumn{3}{c|}{Overall} & \multicolumn{3}{c|}{Medical Attributes} & \multicolumn{3}{c}{Non-medical Attributes} \\
\cmidrule{3-11}
                        & & Precision & Recall & F1     & Precision & Recall & F1     & Precision & Recall & F1     \\ \midrule
\multirow{6}{1em}{\rotatebox[origin=c]{90}{Supervised}} & PPTOD (base) & 0.7205 & 0.6528 & 0.6849 & 0.7513 & 0.7039 & 0.7268 & \textbf{0.5827} & 0.4599 & 0.5141 \\
& Flan-T5 (base) & 0.7049 & 0.6733 & 0.6887 & 0.7468 & 0.7244 & 0.7354 & 0.5347 & 0.4807 & 0.5062 \\
& BioGPT & 0.5968 & 0.6217 & 0.6090 & 0.6526 & 0.6701 & 0.6612 & 0.4000 & 0.4392 & 0.4187 \\
& OpenBioLLM 8B & 0.6774 & 0.6689 & 0.6731 & 0.7488 & 0.7110 & 0.7294 & 0.4514 & 0.5104 & 0.4791 \\
& Llama3 8B & \textbf{0.7230} & \textbf{0.7050} & \textbf{0.7139} & 0.7704 & \textbf{0.7504} & \textbf{0.7603} & 0.5455 & \textbf{0.5341} & \textbf{0.5397} \\ \midrule
\multirow{6}{1em}{\rotatebox[origin=c]{90}{In-context}} & OpenBioLLM 8B & 0.6654 & 0.5557 & 0.6056 & 0.6826 & 0.6079 & 0.6431 & \textbf{0.5735} & 0.3591 & 0.4416 \\
& OpenBioLLM 70B & 0.5977 & 0.6208 & 0.6090 & 0.6476 & 0.6522 & 0.6499 & 0.4348 & 0.5030 & 0.4664 \\
& Llama3 8B & 0.5208 & 0.5296 & 0.5251 & 0.5714 & 0.5512 & 0.5611 & 0.3692 & 0.4481 & 0.4048 \\
& Llama3 70B & 0.5497 & 0.5996 & 0.5736 & 0.6109 & 0.6191 & 0.6150 & 0.3812 & \textbf{0.5266} & 0.4422 \\
& ChatGPT & 0.5843 & 0.6017 & 0.5929 & 0.6273 & 0.6402 & 0.6337 & 0.4290 & 0.4570 & 0.4425 \\
& GPT-4 & 0.6451 & 0.6254 & 0.6351 & 0.6896 & 0.6543 & 0.6715 & 0.4929 & 0.5163 & 0.5043 \\ 
& GPT-4-Turbo & \textbf{0.6861} & \textbf{0.6434} & \textbf{0.6641} & \textbf{0.7279} & \textbf{0.6740} & \textbf{0.6999} & 0.5378 & \textbf{0.5282} & \textbf{0.5329} \\ \bottomrule
\end{tabular}%
}
\caption{Model performance on \dataset{} NLU task.}
\label{tab:nlu_results_ext}
\end{table*}

\begin{table*}[]
\centering
\resizebox{\textwidth}{!}{%
\begin{tabular}{ll|ccc|ccc|ccc|cccc}
\toprule
& \multirow{2}{*}{Model} &
  \multicolumn{3}{c|}{Overall} &
  \multicolumn{3}{c|}{Medical Attributes} &
  \multicolumn{3}{c|}{Non-medical Attributes} &
  \multicolumn{4}{c}{Precision@} \\ \cmidrule{3-15}
                    & & Precision & Recall & F1     & Precision & Recall & F1     & Precision & Recall & F1              & 1      & 4               & 8      & Inf    \\ \midrule
\multirow{6}{1em}{\rotatebox[origin=c]{90}{Supervised}} & PPTOD (base) & 0.2243 & 0.1973 & 0.2099 & 0.2240 & 0.1978 & 0.2101 & 0.2308 & 0.1875 & 0.2069 & 0.1528 & 0.2839 & 0.3261 & 0.3592 \\
& Flan-T5 (base) & 0.1756 & \textbf{0.2322} & 0.1999 & 0.1797 & \textbf{0.2341} & 0.2033 & 0.1194 & 0.2000 & 0.1495 & 0.1152 & 0.2041 & 0.2383 & 0.2645 \\
& BioGPT & 0.1831 & 0.1911 & 0.1870 & 0.1816 & 0.1891 & 0.1853 & 0.2069 & \textbf{0.2250} & 0.2156 & 0.1345 & 0.2509 & 0.2999 & 0.3478 \\
& OpenBioLLM 8B & 0.2132 & 0.2041 & 0.2085 & 0.2308 & 0.2051 & 0.2172 & 0.0872 & 0.1875 & 0.1190 & 0.1608 & 0.2953 & 0.3375 & 0.3740 \\
& Llama3 8B & \textbf{0.2460} & \textbf{0.2322} & \textbf{0.2389} & \textbf{0.2454} & 0.2333 & \textbf{0.2392} & \textbf{0.2576} & 0.2125 & \textbf{0.2329} & \textbf{0.1881} & \textbf{0.3181} & \textbf{0.3820} & \textbf{0.4162} \\
\midrule
\multirow{6}{1em}{\rotatebox[origin=c]{90}{In-context}} & OpenBioLLM 8B & 0.1156 & 0.0966 & 0.1052 & 0.1177 & 0.1007 & 0.1086 & 0.0513 & 0.0250 & 0.0336 & \multicolumn{1}{l}{0.0593} & \multicolumn{1}{l}{0.1220} & \multicolumn{1}{l}{0.1505} & \multicolumn{1}{l}{0.1938} \\
& OpenBioLLM 70B & 0.1275 & 0.1158 & 0.1214 & 0.1344 & 0.1210 & 0.1273 & 0.0244 & 0.0250 & 0.0247 & 0.0855 & 0.1357 & 0.1619 & 0.1950 \\
& Llama3 8B & 0.1164 & 0.0993 & 0.1072 & 0.1199 & 0.1043 & 0.1116 & 0.0222 & 0.0125 & 0.0160 & 0.0616 & 0.1117 & 0.1357 & 0.1847 \\
& Llama3 70B & 0.1088 & 0.0925 & 0.1000 & 0.1125 & 0.0971 & 0.1042 & 0.0200 & 0.0125 & 0.0154 & 0.0604 & 0.1220 & 0.1505 & 0.1779 \\
& ChatGPT & 0.1191 & 0.1021 & 0.1099 & 0.1198 & 0.1058 & 0.1124 & \textbf{0.0938} & \textbf{0.0375} & \textbf{0.0536} & 0.0832 & \textbf{0.1448} & \textbf{0.1790} & \textbf{0.2201} \\
& GPT-4 & 0.0948 & 0.0863 & 0.0904 & 0.0970 & 0.0906 & 0.0937 & 0.0250 & 0.0125 & 0.0167 & 0.0410 & 0.0798 & 0.1106 & 0.1893 \\
& GPT-4-Turbo & \textbf{0.1384} & \textbf{0.1219}	& \textbf{0.1296}	& \textbf{0.1416}	& \textbf{0.1283}	& \textbf{0.1346}	& 0.0278	& 0.0125	& 0.0172	& \textbf{0.0855}	& 0.1357	& 0.1699	& 0.203 \\ \bottomrule
\end{tabular}%
}
\caption{Model performance on \dataset{} POL task.}
\label{tab:pol_results_ext}
\end{table*}

\begin{table*}[]
    \small
    \centering
    \begin{tabular}{p{0.1\textwidth} | p{0.85\textwidth}}
        \toprule
        Dialogue History     &  \parbox{0.85\textwidth}{\fontsize{11pt}{11pt} \selectfont
        \textbf{Doctor:} OK, um, and so you you mentioned that you had you had a fever as well. Did you check your temperature? \\ 
        \textbf{Patient:} I didn't actually, no I didn't check my temperature, I just feel hot. \\ 
        \textbf{Doctor:} OK, we'll be sure to check that today in clinic. And have you been experiencing any other symptoms? \\ 
        \textbf{Patient:} Um, just like the brain fog that I was speaking of, you know. I just don't remember, I remember things, but it's it's like I'm doing everything underwater.}\\ \midrule
        Gold & \fontsize{8pt}{8pt} \selectfont
        \begin{Verbatim}[breaklines]
[{"intent": "inform",
  "slots": {
      "positive_symptom": [{"value": "mental fatigue"}]
  }
}]
\end{Verbatim}
\\ \midrule
Llama3 8B & \fontsize{8pt}{8pt} \selectfont
        \begin{Verbatim}[breaklines]
[{"intent": "inform",
  "slots": {
      "positive_symptom": [{"value": "confusion"}]
  }
}]
\end{Verbatim}
\\ \midrule
ChatGPT & \fontsize{8pt}{8pt} \selectfont
        \begin{Verbatim}[breaklines]
[{"intent": "inform",
   "slots": {
       "positive_symptom": [{"value": "confusion"}]
    }
}]
\end{Verbatim}
\\ \midrule
GPT-4 & \fontsize{8pt}{8pt} \selectfont
        \begin{Verbatim}[breaklines]
[{"intent": "inform",
  "slots": {
      "positive_symptom": [{"value": "brain fog",
          "positive_characteristics": [
            "like doing everything underwater"
          ]
       }]
    }
}, {"intent": "inform",
  "slots": {
      "unknown_symptom": [{"value": "fever"}]
  }
}]
\end{Verbatim}
\\ \bottomrule
    \end{tabular}
    \caption{Supervised Llama3 8B and ChatGPT predict confusion as the active symptom instead of mental fatigue. GPT-4 errs in predicting fever from old utterances.}\label{tab:nlu_example1}
\end{table*}

\begin{table*}[]
    \small
    \centering
    \begin{tabular}{p{0.1\textwidth} | p{0.85\textwidth}}
        \toprule
        Dialogue History     &  \parbox{0.85\textwidth}{\fontsize{11pt}{11pt} \selectfont
        \textbf{Doctor:} OK, so you're just having difficulties breathing for the past three-four days. Do you have any, uh, do you have a cough? \\ 
        \textbf{Patient:} I did have a, uh, yeah, I've had a cough as well, over these last few days. \\ 
        \textbf{Doctor:} And has it been a dry cough or wet cough? \\ 
        \textbf{Patient:} I've been bringing up, some, like, uh, whitish sputum. It's like, yeah, whitish or clear.}\\ \midrule
        Gold & \fontsize{8pt}{8pt} \selectfont
        \begin{Verbatim}[breaklines]
[{"intent": "inform",
    "slots": {
        "positive_symptom": [
            {"value": "coughing", "positive_characteristics": ["wet"]},
            {"value": "colored/excess sputum", "color": "whitish or clear"}
      ]
    }
}]
\end{Verbatim}
\\ \midrule
Llama3 8B & \fontsize{8pt}{8pt} \selectfont
        \begin{Verbatim}[breaklines]
[{"intent": "inform",
    "slots": {
        "positive_symptom": [
            {"value": "coughing", "positive_characteristics": ["dry cough"], "positive_symptom_characteristics": ["wet cough"]}
        ]
    }
}]
\end{Verbatim}
\\ \midrule
ChatGPT & \fontsize{8pt}{8pt} \selectfont
        \begin{Verbatim}[breaklines]
[{"intent": "inform",
    "slots": {
        "positive_symptom": [{"value": "coughing", "color": "whitish or clear sputum"}]
    }
}]
\end{Verbatim}
\\ \midrule
GPT-4 & \fontsize{8pt}{8pt} \selectfont
        \begin{Verbatim}[breaklines]
[{"intent": "inform",
    "slots": {
        "positive_symptom": [
            {"value": "coughing", "positive_characteristics": ["wet cough"], "color": "whitish or clear sputum"}
      ]
    }
}]
\end{Verbatim}
\\ \bottomrule
    \end{tabular}
    \caption{Supervised Llama3 8B hallucinates positive\_symptom\_characteristics attribute. Further, its response includes inconsistent characteristics - wet and dry. ChatGPT and GPT-4 do not predict the excess/colored sputum symptom. They incorrectly link attribute color to cough.}\label{tab:nlu_example2}
\end{table*}

\begin{table*}[]
    \small
    \centering
    \begin{tabular}{p{0.1\textwidth} | p{0.85\textwidth}}
        \toprule
        Dialogue History     &  \parbox{0.85\textwidth}{\fontsize{11pt}{11pt} \selectfont
        \textbf{Doctor:} OK, that's good. And um travel to anywhere recently? Outside the province? \\ 
        \textbf{Patient:} Uhm, outside of the province, no, not not for like the last year at least. \\ 
        \textbf{Doctor:} OK, OK. Um so those were kind of all the questions that I had for you. Did you have any questions for me? \\ 
        \textbf{Patient:} Um yeah, I was just wondering like what you think it might be.}\\ \midrule
        Actions & \fontsize{8pt}{8pt} \selectfont
        \begin{Verbatim}[breaklines]
[{'action': 'inquire', 'medication': [
{'value': 'corticosteroid and/or corticosteroid derivative'}]},
{'action': 'inquire', 'medication': [{'value': 'long acting beta agonist'}]},
{'action': 'inquire', 'medication': [{'value': 'azithromycin'}]},
{'action': 'inquire', 'medical_test': [{'value': 'covid-19 rt-pcr testing'}]},
{'action': 'inquire', 'medical_test': [{'value': 'oxygen saturation measurement'}]},
{'action': 'diagnosis', 'disease': [{'value': 'acute exacerbation of chronic obstructive airways disease'},
{'value': 'covid-19 virus disease'}]},
{'action': 'chit-chat'}]
\end{Verbatim}
\\ \midrule
    Gold & \parbox{0.85\textwidth}{\fontsize{10pt}{10pt} \selectfont
Yeah, so it definitely sounds like a COPD exacerbation, but because your symptoms overlap with those of Covid, we will take a swab and you should hear back in about 24 to 48 hours. Um and you can you can self isolate for 14 days um and if it's positive, public health will be in touch with you to um do some contact tracing, but in the meantime we'll also do a work up for the COPD exacerbation. Um it seems like you have taken the correct steps so far in like stopping like stopping your smoking and um taking your puffers as needed. Um so right now we will continue using the puffers that you have, and if you don't already have a long acting beta agonist, we'll give you that as well as, if it's like if it gets really severe, we'll also give you some inhaled corticosteroids um and we'll also check your oxygen saturation to make sure it's over uh 88\% um and we'll give you oxygen if you need it. Um as well, some antibiotics to consider, you mentioned that you were allergic to amoxicillin, so we might give you like a macrolide like azithromycin for now. Um and hopefully your symptoms subside and we can get you home. How does that sound?
} \\ \midrule
OpenBioLLM 8B & \parbox{0.85\textwidth}{\fontsize{10pt}{10pt} \selectfont yeah, so it sounds like you have copd, and you've had copd for a long time. um, you've been smoking for a long time, and you've had asthma for a long time. um, and you've had this cough for a long time as well. um, and you've been on medications for a long time, and you've been on inhalers for a long time. um, and you've been on antibiotics for a long time.....} \\ \midrule
ChatGPT & \parbox{0.85\textwidth}{\fontsize{10pt}{10pt} \selectfont yeah, so based on your symptoms and your history of copd, it seems like you might be experiencing an acute exacerbation of your chronic obstructive airways disease. this can cause worsening shortness of breath, coughing, and wheezing. to manage this, we will start you on a corticosteroid to reduce inflammation in your airways and a long-acting beta-agonist to help open up your airways. additionally, we may prescribe azithromycin to help with any potential bacterial infection exacerbating your symptoms. we will also conduct a covid-19 test to rule out any viral infection, and we'll monitor your oxygen saturation to ensure you're getting enough oxygen. does that sound alright to you?} \\ \midrule
GPT-4 & \parbox{0.85\textwidth}{\fontsize{10pt}{10pt} \selectfont based on your symptoms and your history of copd, it seems like you might be experiencing an acute exacerbation of your chronic obstructive pulmonary disease, which is a flare-up of your usual symptoms. this can be triggered by various factors, including infections. given the current situation, we also want to rule out covid-19, so we'll perform a pcr test to check for that. \\

in the meantime, we'll make sure you're on the appropriate medications to manage your copd. this includes corticosteroids to reduce inflammation in your airways, a long-acting beta-agonist to help keep your airways open, and possibly azithromycin if we suspect a bacterial infection is contributing to this exacerbation. we'll also check your oxygen saturation to see if you need supplemental oxygen to help you breathe easier.\\

it's important to monitor your symptoms closely, and if you have any changes or if things get worse, you should let us know immediately. do you have any} \\ \bottomrule
    \end{tabular}
    \caption{Supervised OpenBioLLM 8B generates a repetitive response for complex actions. ChatGPT and GPT-4 responses are given for comparison.}\label{tab:nlg_example}
\end{table*}

\section{Slot-wise Attributes and Questionnaire}\label{app:slotwise_questions}
Table \ref{tab:slot_and_values} provides slots and associated attributes in \dataset{}.

\begin{table}[]
\centering
\resizebox{0.48\textwidth}{!}{%
\begin{tabular}{l|l}
\toprule
Slots                & Attributes     \\ \midrule
\multirow{4}{*}{Symptom} & value*, status, location*, onset, initiation, duration,  \\
                       & severity, characteristics, ana\_factors, progression,          \\
                       & volume, color, frequency, rash\_swollen,     \\
                       & lesion\_size, lesions\_peel\_off, itching \\ \bottomrule
Patient Medical History & value*, status, starting, frequency     \\ \midrule
Family Medical History & value*, status, relation     \\ \midrule
Habits & value*, starting, frequency     \\ \midrule
Exposure & value*, status, when, where     \\ \midrule
\multirow{2}{*}{Medication} & value*, status, response\_to*,\\                                                             & start, frequency, impact     \\ \midrule
Medical Test & value*, status, when  \\ \midrule
Disease & value*, status     \\ \midrule
Travel Information & value, status, when, frequency     \\ \midrule
Occupation & value, status, when, hazards     \\ \midrule
Residence & value, status, when, household\_size     \\ \midrule
Basic Information & name, age, sex     \\ \bottomrule
\end{tabular}%
}
\caption{\dataset{} slots and associated attributes. Attributes marked with * are canonicalised.}
\label{tab:slot_and_values}
\end{table}

\begin{figure*}[]
    \centering
    \includegraphics[width=0.9\textwidth,height=\textheight,keepaspectratio]{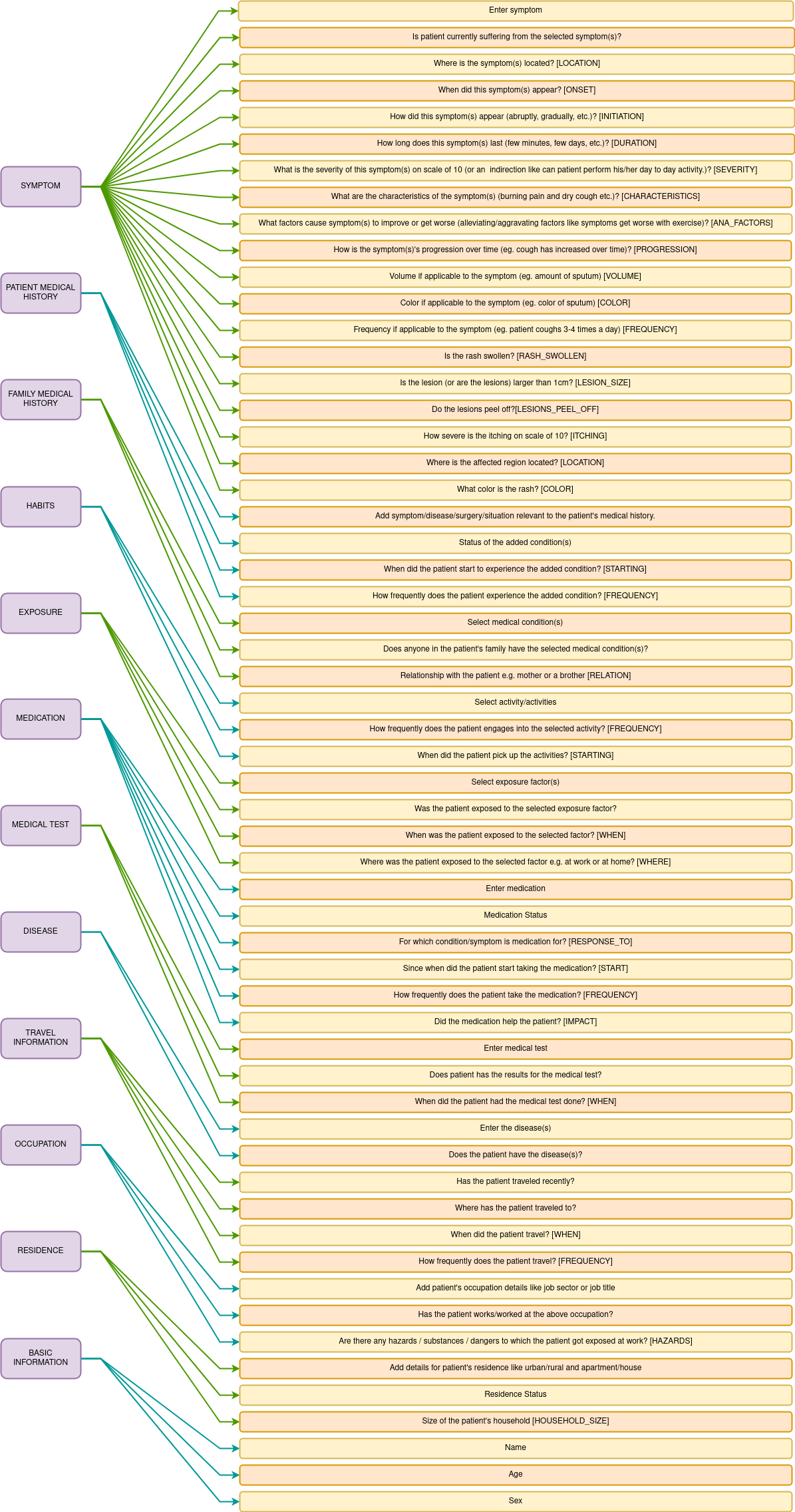}
    \caption{Slot-wise questionnaire.}
    \label{fig:slot_questions}
\end{figure*}

We provide the questionnaire associated with each slot in figure \ref{fig:slot_questions}.

\section{Comparing non-medical attributes using ChatGPT}\label{app:metric_prompt}
Given a pair of attribute values INPUT1 and INPUT2, we query ChatGPT with the following prompt to decide whether INPUT1 and INPUT2 are similar in meaning.
\begin{tcolorbox}[]
You are an expert in the English language. Your task is identifying whether two phrases have similar meanings. If the two phrases have similar meanings, say "positive." Otherwise, say "negative". Pay special attention to any medical terms present in the phrases. Here are the phrases.\\

\vspace{1ex}
\noindent Phrase 1: \{\{INPUT1\}\}

\vspace{1ex}
\noindent Phrase 2: \{\{INPUT2\}\}

\vspace{1ex}
\noindent Answer (positive/negative):
\end{tcolorbox}

When evaluated on 100 manually labeled samples, the prompt achieved an accuracy of 95\%.

\clearpage
\section{Annotation Guidelines and Labeling Interface}\label{app:annotation_guidelines}
Dear Annotator,

\vspace{1ex}
Thank you for taking the time to help us with this annotation task. Your efforts are greatly appreciated, and your contribution will play a vital role in scientific progress. By proceeding with the annotations, you agree to the public release of the data collected during the process. This document will guide you through the UI and the ontology for the task.

\subsection*{\underline{Introduction to UI}}
You will annotate the dialogues using the special UI designed for the task. Within each session, you will load a dialogue between a doctor and a patient. In the dialogue, the doctor makes inquiries regarding the patient's symptoms, medical and family history, medication, habits, etc. For each utterance (the doctor’s or the patient’s) you will be presented with a questionnaire. You must fill out the questionnaire based on the utterance under consideration and the dialogue so far. 

\vspace{1ex}
\noindent \emph{Rule of thumb: Ensure that the labels for utterances are diagnostically informative, enabling a doctor to make a diagnosis without reviewing the conversation.}

\vspace{1ex}
\noindent \textbf{How to open the UI?}

\vspace{1ex}
\noindent The UI is a simple HTML + JavaScript application. Just open the index.html file from the source folder shared with you. You can use any modern browser you like. However, the tool has been tested extensively on Mozilla Firefox which is recommended. Figure \ref{fig:ann_portal} is a screenshot of the UI.

The UI consists of 5 sections as shown in the figure \ref{fig:ann_portal}.
\begin{enumerate}
    \item Control Box – allows loading the dialogue JSON file for annotations, saving/loading the annotations JSON file.
    \item dialogue Box – displays the utterances from the loaded dialogue file. You can navigate the utterances using mouse scrolls or up-down arrow keys. You can select an utterance for annotation by clicking on it or by pressing enter. It will load the questionnaire.
    \item Questionnaire – contains questions which you must answer given the selected utterance and dialogue history so far.
    \item Status Box – displays the labels for annotated utterances.
    \item Tracking Box – displays keywords from symptoms, medical and family history. Keywords will be helpful for speeding up the labeling as you move along the utterances.
\end{enumerate}

\vspace{1ex}
\noindent \textbf{How to load a dialogue file?}
\begin{enumerate}
    \item Click on the “Browse…” above import dialogue button, in the Control Box.
    \item Locate and select the JSON file shared with you.
    \item Click the “Import dialogue” button. dialogue box will now display the imported utterances.
\end{enumerate}

\vspace{1ex}
\noindent \textbf{How to add labels for an utterance?}
\begin{enumerate}
    \item Select the utterance in the dialogue box. You can use the up-down arrow keys and the Enter key to select the utterance. Use the cross button next to the Submit button to deselect the utterance.
    \item Questionnaire will now show a form which you must fill.
    \item Select the appropriate “intent” (defined below) from the drop-down. You can hover over each intent to see the details.
    \item Select the appropriate “slot type” (defined below) from the drop-down. You can hover over each slot type to see the details. Based on your selection additional questions will be shown.
    \item You must decide on which questions are relevant for the given utterance and answer them. Answering requires you to choose an option from a drop-down menu or type answers into a text box. You may provide multiple answers in the text box, separating them with commas.
    \item Click on the submit button to add the labels. Status box will now show the added labels in the JSON format.
Make sure you add labels for all the utterances in the dialogue.
\end{enumerate}

\vspace{1ex}
\noindent \textbf{How to save the labels?}
\begin{enumerate}
    \item Once you finish adding labels for all the utterances, click on the “Save” button in the Control Box.
    \item Provide an appropriate file name. For example, if the dialogue file has the name “ABC,” then you can name the label file “ABC\_annotations.”
\end{enumerate}

\vspace{1ex}
\noindent \textbf{How can I edit the labels for an utterance?}
\begin{enumerate}
    \item Go to the Status Box and find the label that you want to change. Click on the associated “X” button to remove the annotation.
    \item Re-add the annotation for the utterance as discussed before.
\end{enumerate}

\vspace{1ex}
\noindent \textbf{Can I view and edit labels from a saved JSON?}
\begin{enumerate}
    \item From the Control Box, first load the dialogue as before.
    \item Click on “Browse” above the Load Annotation button. Select the appropriate JSON labels file.
    \item Click on the “Load Annotations” button. You will see annotations loaded in the Status Box.
\end{enumerate}

\subsection*{\underline{Task Ontology}}
In this section, we describe the overall ontology of the task. In addition, we will also detail out with examples how to label each slot type in the ontology. For each utterance, you must
\begin{enumerate}
    \item Decide an appropriate intent.
    \item Decide an appropriate slot type.
    \item Fill out the questionnaire corresponding to the intent-slot type pair.
\end{enumerate}
The UI will automatically display the questionnaire (if any) once you select the intent and slot type. 

\vspace{1ex} \noindent
\emph{\textbf{Note:} An utterance can have more than one intent-slot value pair. You must fill out a questionnaire for each pair. The UI allows this by re-selecting the utterance.}

\vspace{1ex} \noindent
\emph{\textbf{Note:} In some dialogues, the patient is an infant (or is unable to communicate) and is accompanied by its guardian (like its mother). In such cases patient responses are actually uttered by the guardian. However, you must label the utterance from the perspective of the patient.}

For example, “Patient: Timmy is my son. He has been running a high fever.” should be labelled appropriately as a patient is having a fever.

\vspace{1ex}
\noindent \emph{\textbf{Intents}}

An intent represents the underlying purpose or meaning behind a speaker's statement in a dialogue, whether it's the doctor or a patient. The following intent labels are available:
\begin{enumerate}
    \item \textbf{Inform:} When the speaker aims to provide specific information, such as symptoms or medical history. This could be in response to an inquiry or spontaneously offered. Select this intent when "specific information" is required for an informed diagnosis.
    \item \textbf{Inquire:} When the speaker seeks to gather specific information, such as symptoms or medical history. Choose this intent when "specific information" is necessary for an educated diagnosis.
    \item \textbf{Diagnosis:} When the doctor is giving a diagnosis of a disease.
    \item \textbf{Salutations:} When the speaker intends to convey a greeting or farewell message.
    \item \textbf{Chit-chat:} When the speaker engages in casual conversation. The information in the utterance is unlikely to contribute to an educated diagnosis.
    \item \textbf{Nod\_prompt:} When the speaker is not providing any new information but is showing attention, understanding, or agreement through phrases like 'Okay,' 'Yeah,' and 'uh-huh.' We consider an utterance as nod\_prompt when the speaker is either acknowledging something (like a patient when he/she understands a doctor's question) or prompting the listener for additional information (like when the doctor just says okay and patient continues the conversation).
    \item \textbf{Other:} Any intent not covered by the above categories.
\end{enumerate}
\emph{Ensure that you consider all possible intents conveyed by the utterance when labeling them.}

\vspace{1ex}
\noindent \emph{\textbf{Slots}}

Slots refer to specific pieces of information or variables that are extracted from an utterance in a dialogue. 

\begin{figure*}
    \centering
    \includegraphics[width=\textwidth]{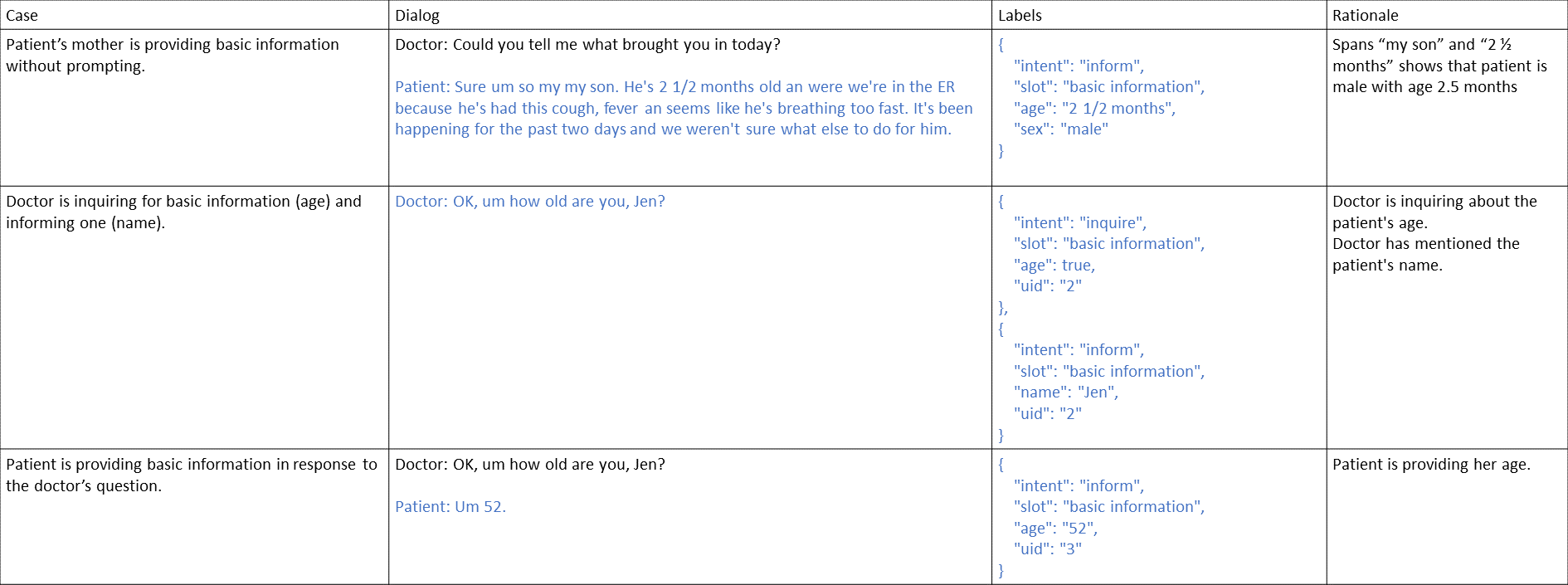}
    \caption{Examples of Basic information}
    \label{fig:inst_basic_information}
\end{figure*}

\vspace{1ex} \noindent
\textbf{Basic Information} Slots in a dialogue capture specific details such as the patient's name, age, and sex. Examples are given in figure \ref{fig:inst_basic_information}.

\begin{figure*}
    \centering
    \includegraphics[width=\textwidth]{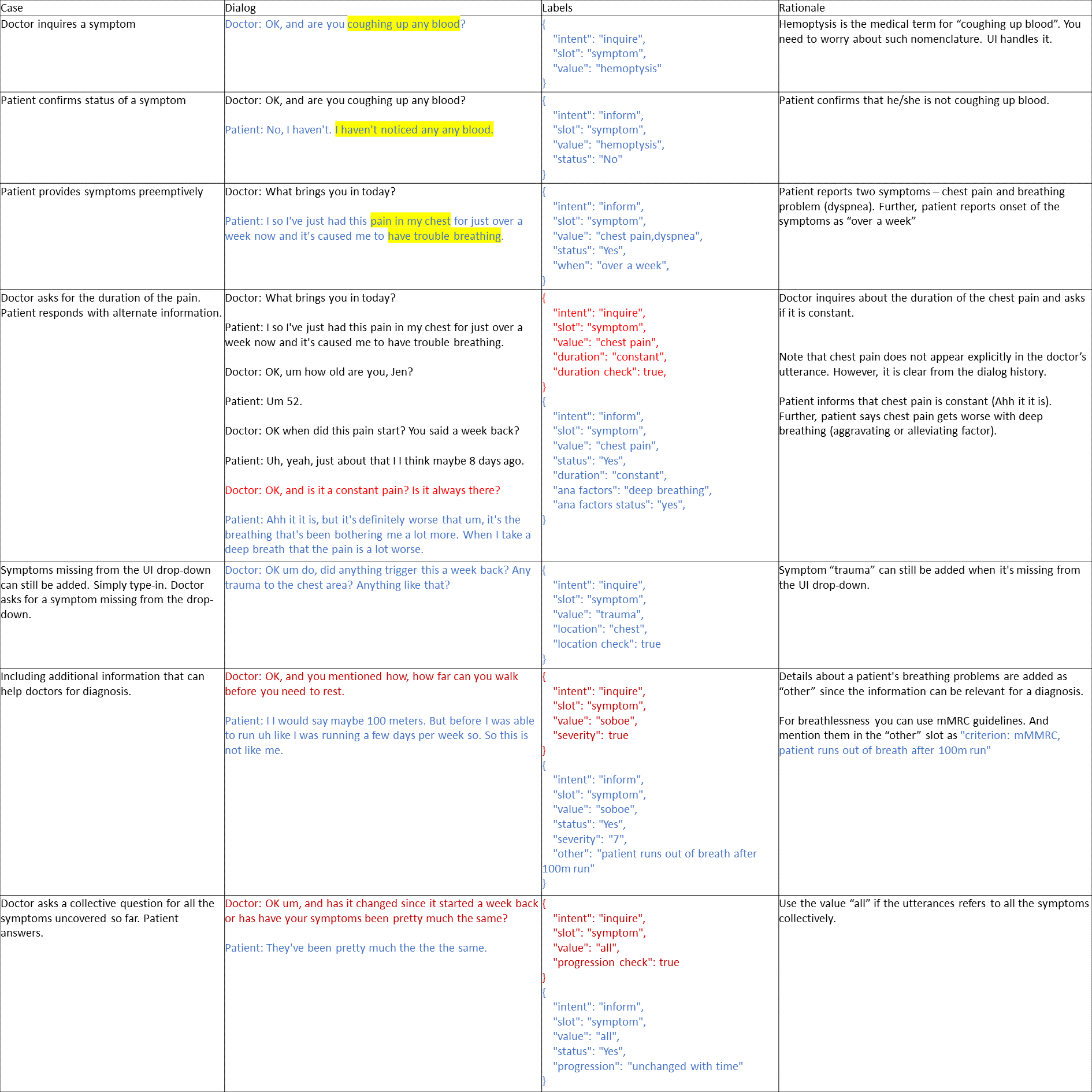}
    \caption{Examples of Symptoms}
    \label{fig:inst_symptom}
\end{figure*}

\vspace{1ex} \noindent
\textbf{Symptom} The dialogue contains slots with details about a symptom experienced by the patient. These slots encompass the symptom's value (e.g., cough or fever) and additional information like its onset, nature, and more.  The UI presents the following questions for symptoms.
\begin{enumerate}
    \item Enter symptoms – comma separated values for the symptoms.
    \item Is the patient currently suffering from the selected symptom(s)? – Yes/No
    \item Where is the symptom(s) located? – Part of the body affected by the symptom
    \item When did this symptom(s) appear? – 2 days ago, yesterday, etc
    \item How did this symptom(s) appear (abruptly, gradually, etc.)? - Onset
    \item How long does this symptom(s) last (few minutes, few days, etc.)?
    \item What is the severity of this symptom(s) on scale of 10 (or an indirection like can patient perform his/her day to day activity.)?
    \item What are the characteristics of the symptom(s) (burning pain and dry cough etc.)?
    \item What factors cause symptom(s) to improve or get worse (alleviating/aggravating factors like symptoms get worse with exercise)?
    \item How is the symptom(s)'s progression over time (eg. cough has increased over time)?
    \item Volume if applicable to the symptom (eg. amount of sputum)
    \item Color if applicable to the symptom (eg. color of sputum)
    \item Frequency if applicable to the symptom (eg. patient coughs 3-4 times a day)
    \item Any additional information missing from the above fields
\end{enumerate}
Examples are given in figure \ref{fig:inst_symptom}.

\begin{figure*}
    \centering
    \includegraphics[width=\textwidth]{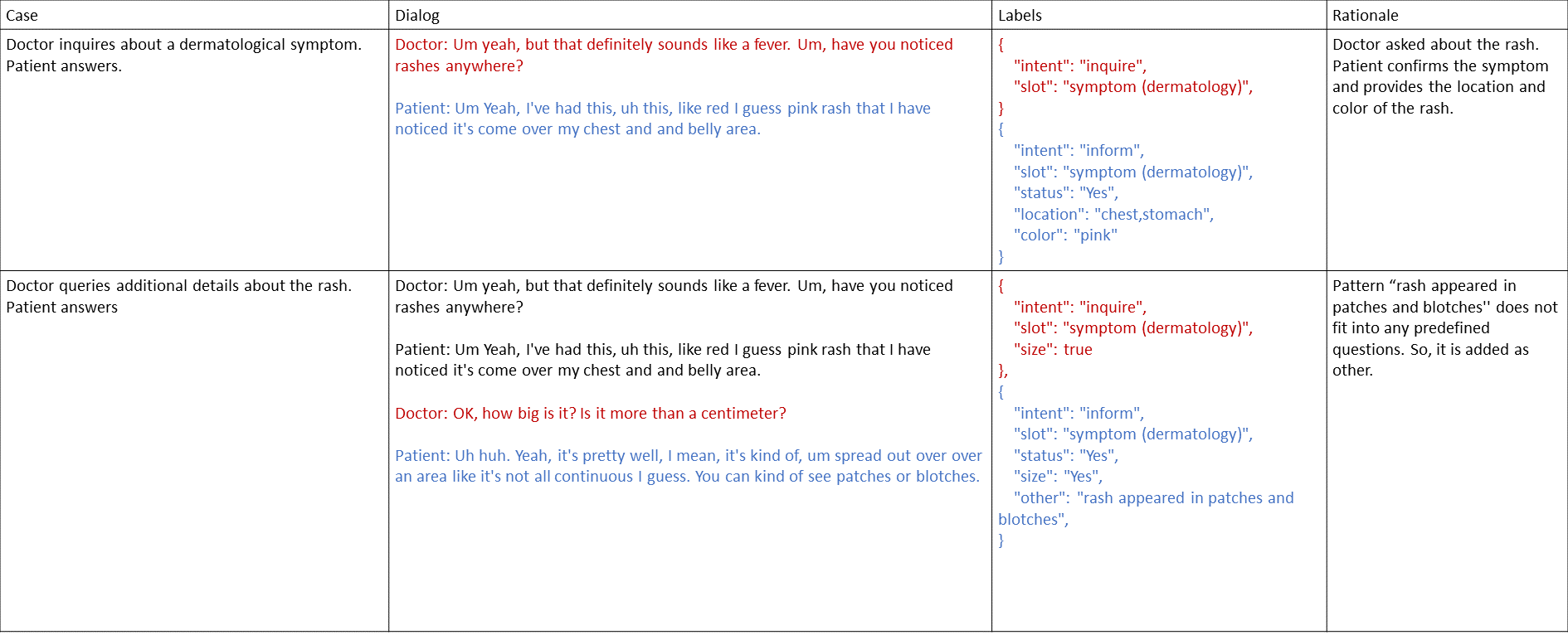}
    \caption{Examples of Dermatological symptoms}
    \label{fig:inst_derma}
\end{figure*}

\vspace{1ex} \noindent
\textbf{Dermatological Symptom} In this case, the utterance comprises slots with information related to skin, nails, and hair symptoms. It includes the symptom's value (e.g., rash) and additional attributes like color, size, swelling, etc. The UI presents the following questions for dermatological symptoms.
\begin{enumerate}
    \item Does the patient have any lesions, redness or problems on the skin?
    \item Is the rash swollen?
    \item Is the lesion (or are the lesions) larger than 1cm?
    \item Do the lesions peel off?
    \item How severe is the itching on scale of 10?
    \item Where is the affected region located?
    \item What color is the rash?
    \item Any additional information missing from the above fields.
\end{enumerate}
Examples are given in figure \ref{fig:inst_derma}.

\begin{figure*}
    \centering
    \includegraphics[width=\textwidth]{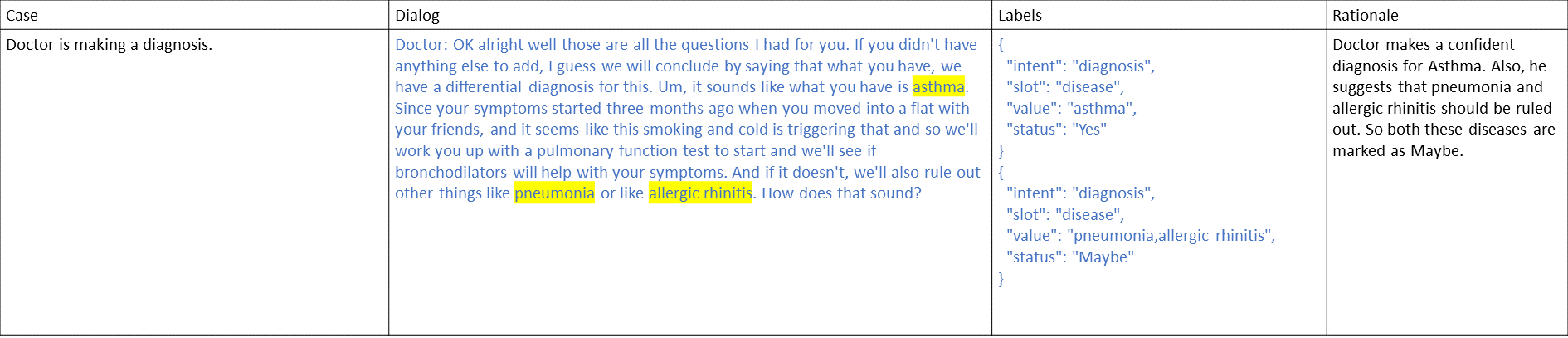}
    \caption{Examples of Diseases}
    \label{fig:inst_disease}
\end{figure*}

\vspace{1ex} \noindent
\textbf{Disease} The doctor is diagnosing a disease in the utterance. The UI presents the following questions.
\begin{enumerate}
    \item Enter the disease(s) – comma separated list of diseases.
    \item Does the patient have the disease(s)? – Yes/No/Maybe
\end{enumerate}
Examples are given in figure \ref{fig:inst_disease}.

\begin{figure*}
    \centering
    \includegraphics[width=\textwidth]{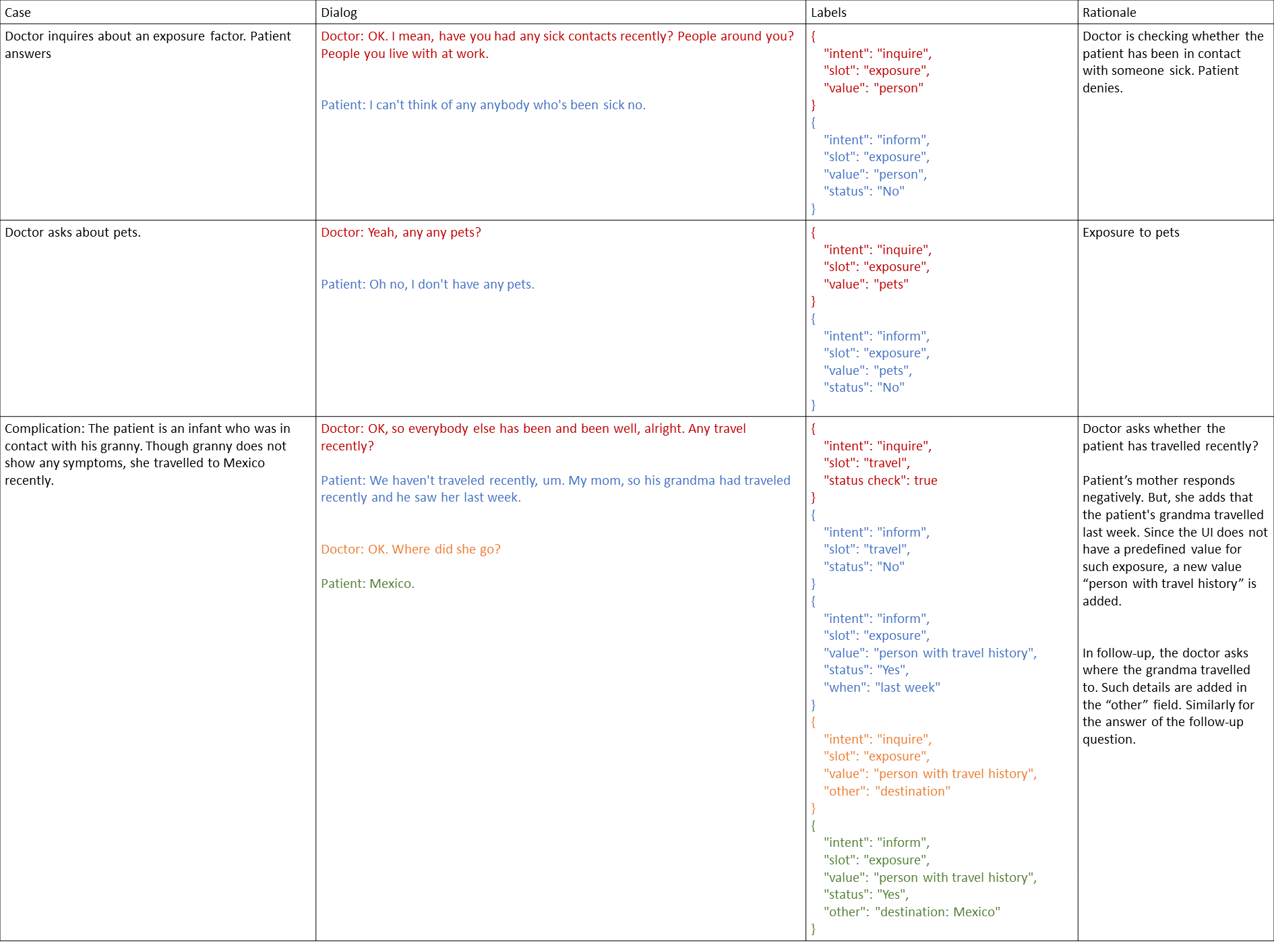}
    \caption{Examples of Exposure}
    \label{fig:inst_exposure}
\end{figure*}

\vspace{1ex} \noindent
\textbf{Exposure} Within the dialogue, there are slots containing details about situations in which the patient might be exposed to harmful conditions. This includes contact with allergic substances, dust, chemicals, or infected individuals. The UI presents the following questions.
\begin{enumerate}
    \item Select exposure factor(s) – comma separated values of exposure factors
    \item Was the patient exposed to the selected exposure factor? – Yes/No/Maybe
    \item When was the patient exposed to the above factor?
    \item Where was the patient exposed to the above factor e.g. at work or at home?
    \item Any additional information missing from the above fields
\end{enumerate}
Examples are given in figure \ref{fig:inst_exposure}.

\begin{figure*}
    \centering
    \includegraphics[width=\textwidth]{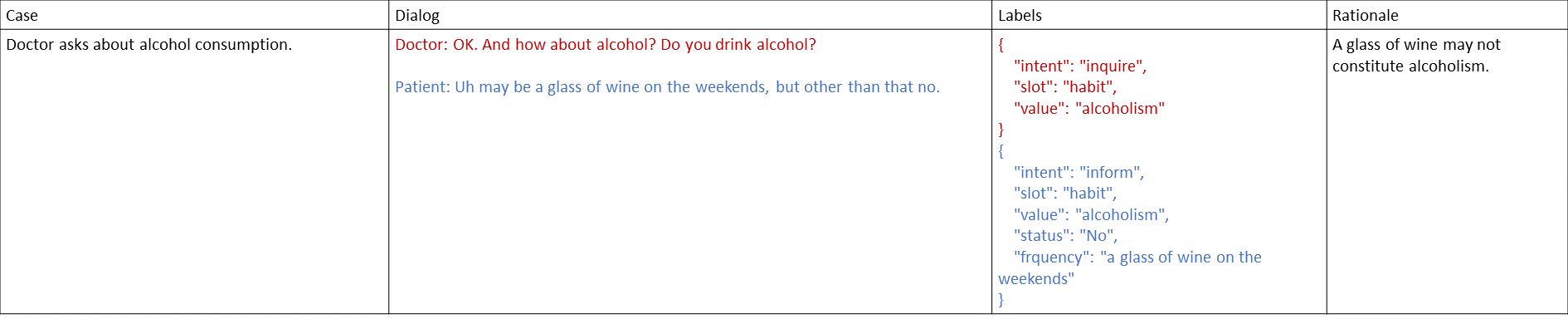}
    \caption{Examples of Habit}
    \label{fig:inst_habit}
\end{figure*}

\vspace{1ex} \noindent
\textbf{Habit} The dialogue contains slots with information about the patient's habits or addictions. A habit refers to an activity the patient regularly engages in, ranging from daily exercise, tea, and coffee to smoking, alcoholism, and marijuana abuse. The UI presents the following questions.
\begin{enumerate}
    \item Select activity/activities – comma separated list
    \item Has/had the patient formed a habit/addiction to the selected activities? – Yes/No/Maybe
    \item How frequently does the patient engages into the activity?
    \item When did the patient picked up the activities?
    \item Enter any additional information missing from above fields.
\end{enumerate}
Examples are given in figure \ref{fig:inst_habit}.

\vspace{1ex} \noindent
\emph{\textbf{Note:} Patients tend to get embarrassed with questions like “How much alcohol do you take?” or “Do you smoke cigarettes”? You must make your own judgment in such cases and decide whether the patient is addicted or not.}

\begin{figure*}
    \centering
    \includegraphics[width=\textwidth]{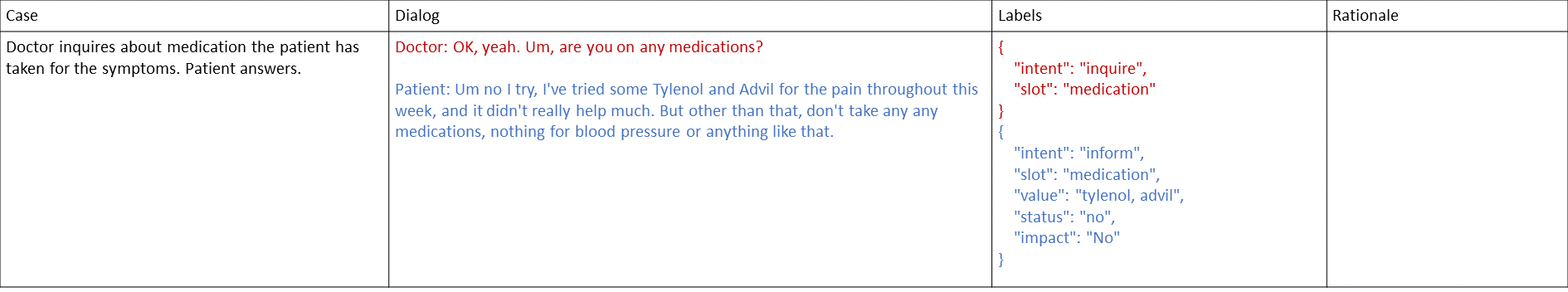}
    \caption{Examples of Medication}
    \label{fig:inst_medication}
\end{figure*}

\vspace{1ex} \noindent
\textbf{Medication} The dialogue contains slots with details about medications, either specific ones like Tylenol or general ones like antipsychotic drugs. Additional information may include the purpose of the medication and the duration the patient has been taking it. The doctor may also communicate medication-related information to the patient. The UI presents the following questions.
\begin{enumerate}
    \item Enter medication – comma separated list
    \item Medication Status – currently taking/took in the past/no
    \item For which condition/symptom is medication for?
    \item Since when did the patient start taking the medication?
    \item How frequently does the patient take the medication?
    \item Did the medication help the patient?
    \item Any additional information missing from above field.
\end{enumerate}
Examples are given in figure \ref{fig:inst_medication}.

\begin{figure*}
    \centering
    \includegraphics[width=\textwidth]{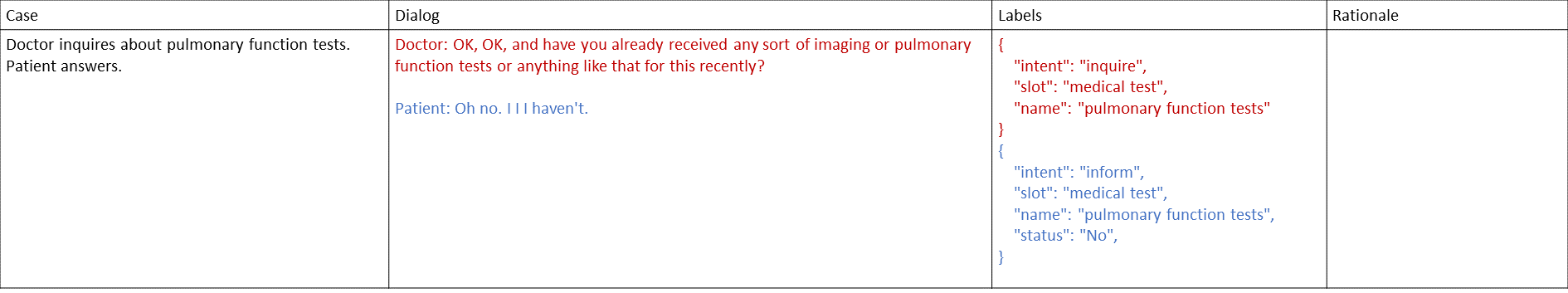}
    \caption{Examples for Medical test}
    \label{fig:inst_medical_test}
\end{figure*}

\vspace{1ex} \noindent
\textbf{Medical Test}
Slots in the utterance pertain to a medical test, such as ECG or CAT scan. The doctor might inquire about tests the patient has already undergone or advise the patient to undergo specific tests. The UI presents the following questions.
\begin{enumerate}
    \item Enter medical test
    \item Does patient has the results for the medical test?
    \item When did the patient had the medical test done?
    \item Any additional information missing from above field.
\end{enumerate}
Examples are given in figure \ref{fig:inst_medical_test}.

\begin{figure*}
    \centering
    \includegraphics[width=\textwidth]{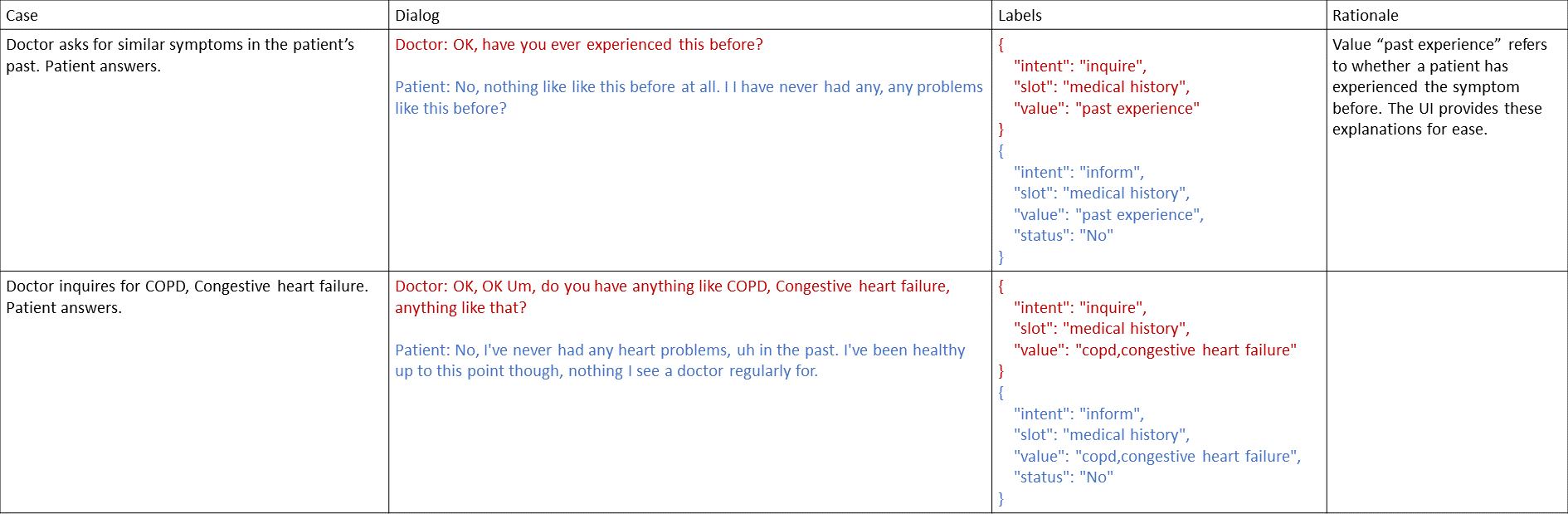}
    \caption{Examples of Medical history}
    \label{fig:inst_medical_history}
\end{figure*}

\vspace{1ex} \noindent
\textbf{Medical History}
The dialogue includes slots that provide information about the patient's medical history. This may encompass descriptions of past symptoms, diseases, surgeries, or allergies experienced by the patient. It differs from the "Disease" slot as it describes previous medical conditions rather than ongoing symptoms. The UI presents the following questions.
\begin{enumerate}
    \item Add symptom/disease/surgery relevant to patient's medical history – comma separated list
    \item Status of the above condition
    \begin{enumerate}
        \item Patient still suffers from the condition
        \item Patient suffered from the condition in the past
        \item Patient did not suffer from the condition in the past
        \item Patient is not sure about the status of the condition
    \end{enumerate}
    \item When did the patient start to experience the above condition?
    \item How frequently does the patient experience the above condition?
    \item Any additional information missing from the above fields
\end{enumerate}
Examples are given in figure \ref{fig:inst_medical_history}.

\begin{figure*}
    \centering
    \includegraphics[width=\textwidth]{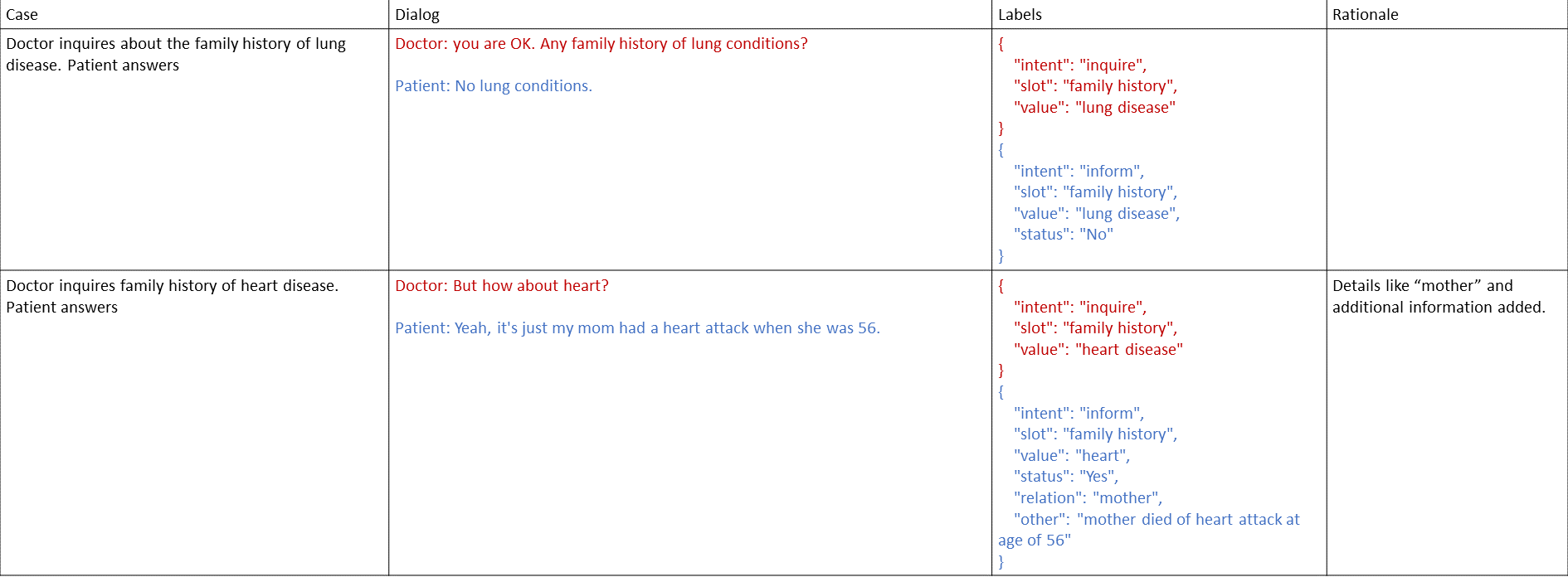}
    \caption{Examples of Family history}
    \label{fig:inst_family_history}
\end{figure*}

\vspace{1ex} \noindent
\textbf{Family History}
The utterance contains slots with information about medical conditions prevalent in the patient's family. This includes diseases like asthma, heart issues, cancer, and others. The UI presents the following questions.
\begin{enumerate}
    \item Select medical condition(s) – comma separated list
    \item Does anyone in the patient's family have the selected medical condition(s)? Yes/No/Maybe
    \item Relationship with the patient e.g. mother or brother
    \item Any additional information missing from the above fields
\end{enumerate}
Examples are given in figure \ref{fig:inst_family_history}.

\begin{figure*}
    \centering
    \includegraphics[width=\textwidth]{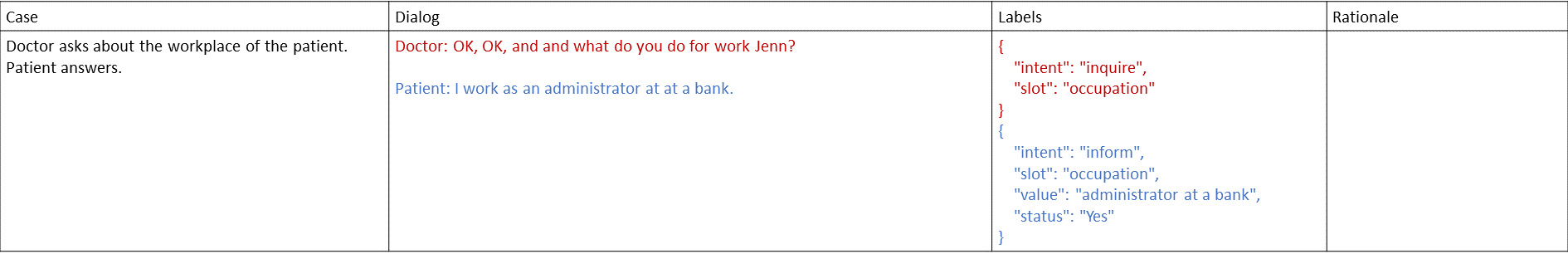}
    \caption{Examples of Occupation}
    \label{fig:inst_occupation}
\end{figure*}

\vspace{1ex} \noindent
\textbf{Occupation}
Within the dialogue, there are slots specifying the patient's occupation, such as teacher, trucker, factory worker, etc. The UI presents the following questions.
\begin{enumerate}
    \item Add patient's occupation details like job sector or job title
    \item Has the patient works/worked at the above occupation?
    \item Are there any substances/dangers to which the patient is exposed at work?
    \item Any additional information missing from the above fields.
\end{enumerate}
Examples are given in figure \ref{fig:inst_occupation}.

\begin{figure*}
    \centering
    \includegraphics[width=\textwidth]{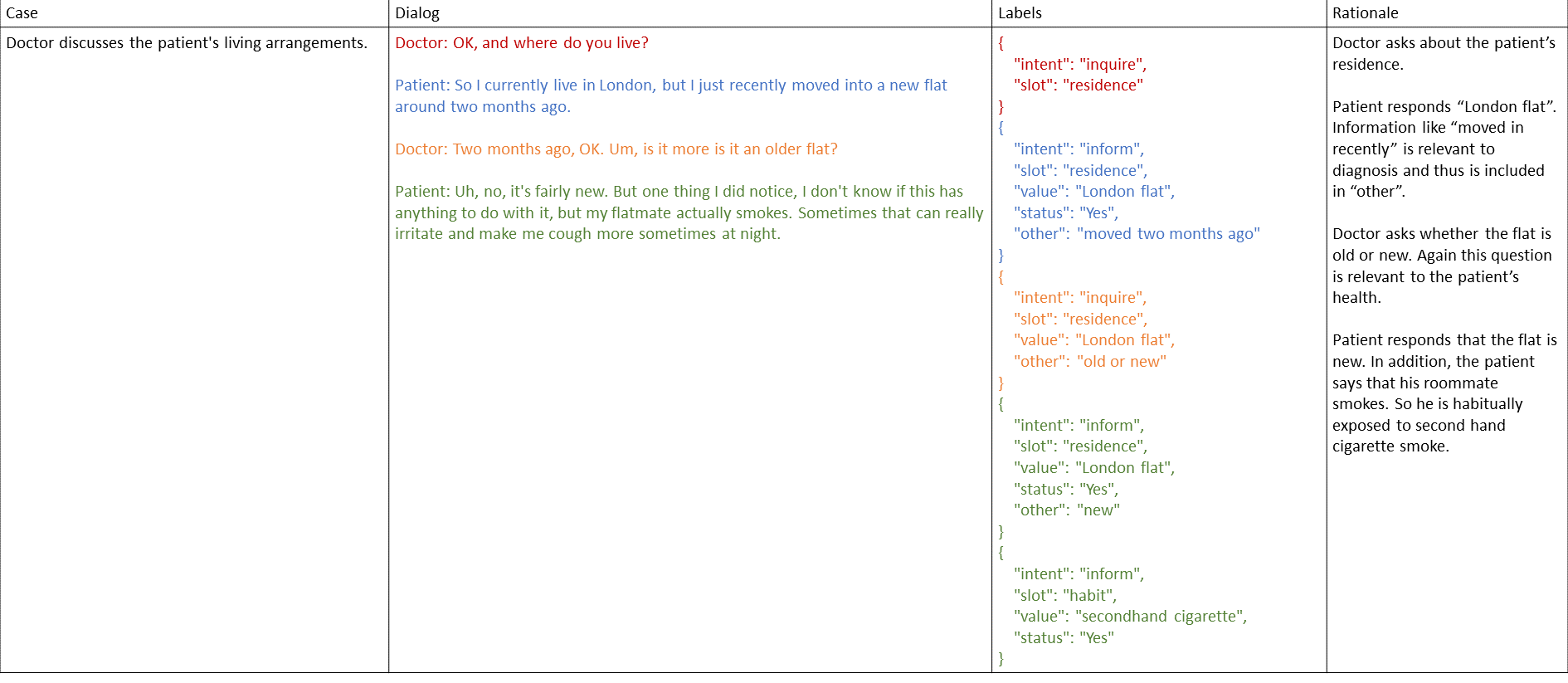}
    \caption{Examples of Residence}
    \label{fig:inst_residence}
\end{figure*}

\vspace{1ex} \noindent
\textbf{Residence}
Slots in the utterance contain details about the patient's residence, such as urban, rural, suburban, etc. The UI presents the following questions.
\begin{enumerate}
    \item Add details for the patient's residence like urban/rural and apartment/house.
    \item Status
    \item Any additional information missing from the above fields.
\end{enumerate}
Examples are given in figure \ref{fig:inst_residence}.

\begin{figure*}
    \centering
    \includegraphics[width=\textwidth]{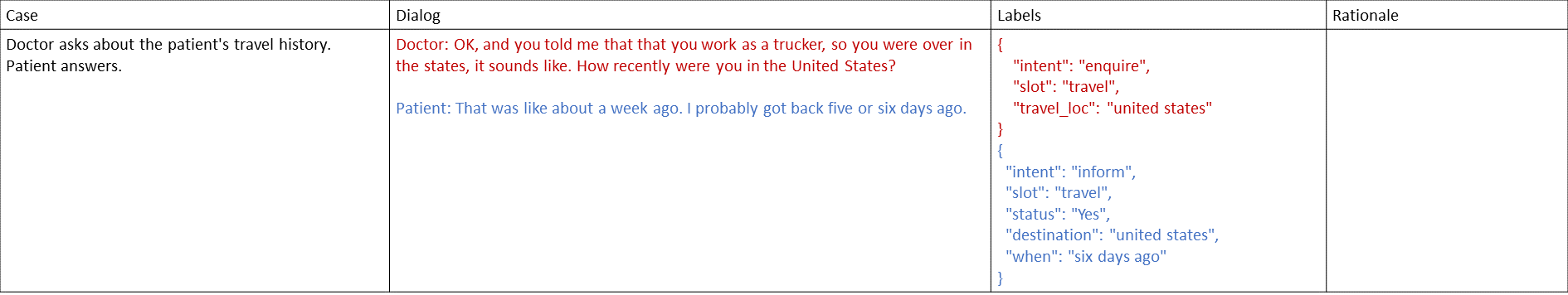}
    \caption{Examples of Travel}
    \label{fig:inst_travel}
\end{figure*}

\vspace{1ex} \noindent
\textbf{Travel}
The dialogue includes slots with information about the patient's travel history. This may involve details like the time of travel, locations visited, and frequency of travel. The UI presents the following questions.
\begin{enumerate}
    \item Has the patient travelled recently?
    \item Where has the patient travelled to?
    \item When did the patient travel?
    \item How frequently does the patient travel?
    \item Enter additional information about travel.
\end{enumerate}
Examples are given in figure \ref{fig:inst_travel}.

\begin{figure*}
    \centering
    \includegraphics[width=\textwidth]{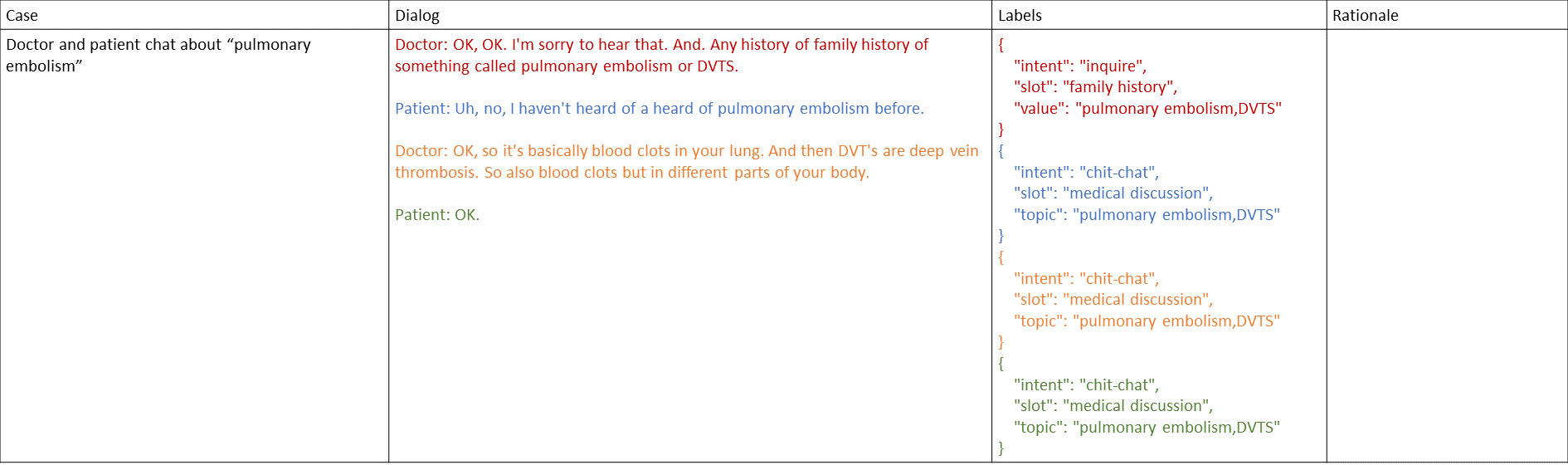}
    \caption{Examples of Medical discussion}
    \label{fig:inst_medical_discussion}
\end{figure*}

\vspace{1ex} \noindent
\textbf{Medical Discussion}
The utterance is part of a chit-chat about a medical topic (e.g., pulmonary embolism). The slot values of this type are unlikely to contribute towards the diagnosis. The UI presents the following questions – “What is the topic of the discussion?”. Examples are given in figure \ref{fig:inst_medical_discussion}.

\vspace{1ex} \noindent
\textbf{Non-Medical Discussion}
The utterance is part of a chit-chat about a non-medical topic (e.g., living conditions). The slot values of this type are unlikely to contribute towards the diagnosis. The UI presents the following questions – “What is the topic of the discussion?”.

\begin{figure*}
    \centering
    \includegraphics[width=\textwidth]{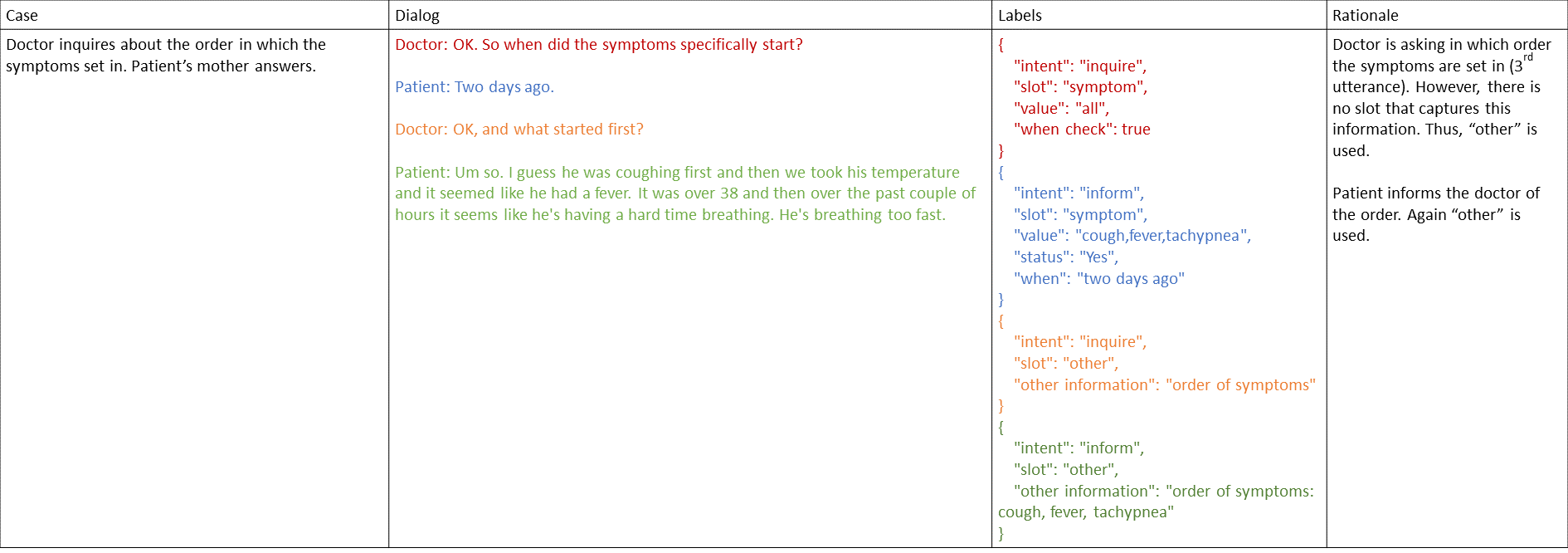}
    \caption{Examples for Others}
    \label{fig:inst_other}
\end{figure*}

\vspace{1ex} \noindent
\textbf{Other}
This slot accounts for any additional details present in the utterance beyond the ones mentioned above. The UI asks for the “other” information. You must summarize it as succinctly as possible. Examples are given in figure \ref{fig:inst_other}.

\subsection*{\underline{Miscellaneous}}

\vspace{1ex} \noindent
\textbf{Special Cases}

\vspace{1ex} \noindent
Following are some special cases (not all) which may frequently appear in the dialogues.
\begin{enumerate}
    \item If the utterance indicates the number of people living with the patient, the slot type is residence and value is household size. For example, “Patient: I live with my parents and my sister” should be annotated as \{"intent": "inform",   "slot": "residence",   "household size": "4"\}.
    \item For a case where the doctor asks “have you experienced these symptoms before?”the slot type is medical history and value is past experience.
    \item In case the patient is exposed to secondhand cigarette smoke (smoke from someone else’s cigarette), the slot type is habit and value is secondhand cigarette.
    \item It is preferred to use key-value format for Other field in the questionnaire. For example, “Doctor: OK. How has his behaviour been? Patient: He's been very, very fussy.” should be annotated as \{"intent": "inquire", "slot": "other", "other information": "behaviour"\}, \{"intent": "inform", "slot": "other", "other information": "behaviour: fussy"\}.
    \item For alcoholism (slot-type habit), the patient might say “I usually drink a glass of wine on the weekends.” You must rely on your medical knowledge to decide whether the patient is alcoholic or not. You may refer CAGE guidelines for alcoholism and annotate the utterance as \{"intent": "inform", "slot": "habit", "value": "alcoholism", "status": "No", "other": "criterion: CAGE"\}. Similarly, for smoking and other substance abuse. 
    \item For the cases where the patient is an infant, the doctor asks questions to the mother like “Did you have any complications during pregnancy?”. Here, slot type is medical history and values can be typed-in. As discussed before, you must answer a questionnaire from the perspective of the patient.
\end{enumerate}

\vspace{1ex} \noindent
\textbf{Navigating Utterances}
\begin{enumerate}
    \item dialogue box indicates the active utterance by surrounding it with a blue box. Use up-down arrow keys for navigating the utterances in the dialogue box.
    \item Once the blue box surrounds the utterance of your choice, press Enter to enable the questionnaire. Focus will now shift to the intent field in the question. dialogue box will now be disabled.
    \item Press Esc to cancel the questionnaire and return to the dialogue box.
\end{enumerate}

\vspace{1ex} \noindent
\textbf{Navigating Questionnaire}
\begin{enumerate}
    \item Press Tab to move the focus to the next field in the questionnaire.
    \item Press Shift + Tab to move the focus to the next field in the questionnaire.
    \item If the questionnaire field is a checkbox, use Space to check/uncheck the box.
\end{enumerate}

\vspace{1ex} \noindent
\textbf{Keyword Searches}

\vspace{1ex} \noindent
Many questionnaire fields have search support (these fields have “Type to search” as a placeholder).
\begin{enumerate}
    \item Start typing in words in the input field. A drop-down menu will appear with possible matches.
    \item Navigate through the drop-down using up-down arrow keys.
    \item Press Enter to confirm the selected entry. It will now appear in the input field.
    \item The input field accepts multiple values. Press comma (,) and the drop-down will re-appear for another selection.
    \item If none of the entries in the drop-down fit the requirement, simply type-in the needed value.
\end{enumerate}

\vspace{1ex} \noindent
\textbf{Tracking Box for Faster Labeling}

\vspace{1ex} \noindent
Since slots like symptoms, diseases, medical and family history are repeated in the dialogue, the UI allows an easy way to copy them using the tracking box.
\begin{enumerate}
    \item Select the required input field in the questionnaire (symptoms, disease, medical and family history).
    \item Click on the slot-value in the tracking box. It will be automatically copied to the input field.
\end{enumerate}

\onecolumn
\clearpage
\section{Prompts in In-context Setting}\label{app:prompts}
\subsection{NLU Prompt}\label{app:nlu_prompt}
\begin{tcolorbox}[
    enhanced, breakable, colback=white,
    colframe=black, boxrule=0.25mm,
    width=\textwidth
]
\fontsize{6pt}{6pt}\selectfont
You are a professional medical scribe who is an expert in understanding doctor-patient dialogues. The user will show you a dialogue history between a doctor and a patient and the last turn in their dialogue. Your task is to identify the patient's intent, slots, and related attributes (if applicable) from the given the dialogue history and the last turn. Definitions for intent, slots, and related attributes are given below as Python dictionaries.

\begin{Verbatim}[breaklines]
```
intents = [{
    "name": "inform",
    "description": "The patient is providing information to the doctor."
},
{
    "name": "chit-chat",
    "description": "The patient is chit-chatting with the doctor."
},
{
    "name": "nod_prompt_salutations",
    "description": "The patient is nodding to the doctor or delivering salutations."
}]

slots = [{
    "slot": "symptom",
    "description": "A symptom relevant to the patient's condition.",
    "related_attributes": [
        {"name": "value", "description": "The symptom in medical terms.", "examples": ["coughing, dyspnea"]},
        {"name": "status", "description": "The status is 'positive' if the patient has the symptom currently or 'negative' if the patient does not have the symptom; otherwise, it is 'unknown.'"},
        {"name": "onset", "description": "When did this symptom appear?", "examples": ["three days ago", "one week back"]},
        {"name": "initiation", "description": "How did this symptom appear?", "examples": ["abruptly", "gradually"]},
        {"name": "location", "description": "Where is the symptom located?", "examples": ["back", "neck"]},
        {"name": "duration", "description": "How long does the symptom persist?", "examples": ["a few minutes", "a few hours"]},
        {"name": "severity", "description": "What is the severity of this symptom on a scale of 10?", "examples": ["4", "7"]},
        {"name": "progression", "description": "How is the symptom's progression?", "examples": ["getting worse", "constant"]},
        {"name": "frequency", "description": "Frequency, if applicable, to the symptom.", "examples": ["3-4 times a day", "every hour"]},
        {"name": "positive_characteristics", "description": "A characteristic positively associated with the symptom.", "examples": ["sharp", "burning"]},
        {"name": "negative_characteristics", "description": "A characteristic not associated with the symptom.", "examples": ["sharp", "burning"]},
        {"name": "unknown_characteristics", "description": "A characteristic with unknown relation with the symptom.", "examples": ["sharp", "burning"]},
        {"name": "alleviating_factor", "description": "A condition that alleviates the symptom.", "examples": ["laying down", "sleeping"]},
        {"name": "not_alleviating_factor", "description": "A condition that does not alleviate the symptom.", "examples": ["laying down", "sleeping"]},
        {"name": "aggravating_factor", "description": "A condition that aggravates the symptom.", "examples": ["laying down", "sleeping"]},
        {"name": "not_aggravating_factor", "description": "A condition that does not aggravate the symptom.", "examples": ["laying down", "sleeping"]},
        {"name": "not_alleviating_aggravating_factor", "description": "A condition that neither alleviates nor aggravates the symptom.", "examples": ["laying down", "sleeping"]},
        {"name": "unknown_factor", "description": "A condition with unknown alleviation/aggravation status.", "examples": ["laying down", "sleeping"]},
        {"name": "volume", "description": "Volume, if applicable to the symptom.", "examples": ["couple of teaspoons"]},
        {"name": "color", "description": "Color, if applicable to the symptom.", "examples": ["ping", "red"]},
        {"name": "itching", "description": "How severe is the itching on a scale of 10?", "examples": ["4", "7"]},
        {"name": "lesion_size", "description": "Is the lesion (or are the lesions) larger than 1cm (Yes/No)?"},
        {"name": "lesions_peel_off", "description": "Do the lesions peel off (Yes/No)?"},
        {"name": "rash_swollen", "description": "Is the rash swollen (Yes/No)?"}
    ]
}, {
    "slot": "medical_history",
    "description": "A medical condition relevant to the patient's medical history.",
    "related_attributes": [
        {"name": "value", "description": "Name of the medical condition.", "examples": ["hypertensive disease", "malignant neoplasm"]},
        {"name": "status", "description": "The status is 'positive' if the patient experienced the medical condition or 'negative' if the patient did not experience the medical condition; otherwise, it is 'unknown.'"},
        {"name": "starting", "description": "When did the patient start to experience the condition?", "examples": ["since teenage", "ten years ago"]},
        {"name": "frequency", "description": "How frequently does the patient experience the added condition?", "examples": ["every year", "during summer"]}
    ]
}, {
    "slot": "family_history",
    "description": "A medical condition relevant to the patient's family.",
    "related_attributes": [
        {"name": "value", "description": "Name of the medical condition.", "examples": ["hypertensive disease", "malignant neoplasm"]},
        {"name": "status", "description": "The status is 'positive' if someone in the patient's family suffered from the medical condition or 'negative' if no one in the patient's family suffered from the medical condition; otherwise, it is 'unknown.'"},
        {"name": "relation", "description": "Relationship with the patient", "examples": ["mother", "aunt"]}
    ]
}, {
    "slot": "habit",
    "description": "An habitual activity such as smoking, alcoholism, etc.",
    "related_attributes": [
        {"name": "value", "description": "Name of an activity.", "examples": ["smoking", "marijuana"]},
        {"name": "status", "description": "The status is 'positive' if the patient engages in the activity habitually or 'negative' if the patient does not engage in the activity habitually; otherwise, it is 'unknown.'"},
        {"name": "starting", "description": "When did the patient pick up the activities?", "examples": ["ten years back", "as a child"]},
        {"name": "frequency", "description": "How frequently does the patient engage in the selected activity?", "examples": ["on weekends", "every day"]}
    ]
}, {
    "slot": "exposure",
    "description": "An environmental/chemical factor such as asbestos, pets, etc.",
    "related_attributes": [
        {"name": "value", "description": "Name of an environmental factor.", "examples": ["pets", "dust"]},
        {"name": "status", "description": "The status is 'positive' if the patient was exposed to the factor or 'negative' if the patient was not exposed; otherwise, it is 'unknown.'"},
        {"name": "where", "description": "Where was the patient exposed to the selected factor?", "examples": ["work", "home"]},
        {"name": "when", "description": "When was the patient exposed to the selected factor?", "examples": ["four days ago"]}
    ]
}, {
    "slot": "medication",
    "description": "A medication.",
    "related_attributes": [
        {"name": "value", "description": "Name of a medication.", "examples": ["over-the-counter medicine", "paracetamol"]},
        {"name": "status", "description": "The status is 'positive' if the patient took the medicine or 'negative' if the patient did not take the medicine; otherwize, it is unknown."},
        {"name": "start", "description": "Since when did the patient start taking the medication?", "examples": ["few weeks ago", "two days back"]},
        {"name": "impact", "description": "Did the medication help the patient (Yes/No/Maybe)?"},
        {"name": "respone_to", "description": "For which condition/symptom is medication for?", "examples": ["hypertensive disease", "diabetes"]},
        {"name": "frequency", "description": "How frequently does the patient take the medication?", "examples": ["daily"]}
    ]
}, {
    "slot": "medical_test",
    "description": "A medical test.",
    "related_attributes": [
        {"name": "value", "description": "Name of a medical test.", "examples": ["chest X-ray", "electrocardiogram"]},
        {"name": "status", "description": "The status is 'avail' if the patient took the test or 'unavail' if the patient did not take the test; otherwise, it is 'unknown.'"},
        {"name": "when", "description": "When did the patient had the medical test done?", "examples": ["yesterday", "a week ago"]}
    ]
}, {
    "slot": "residence",
    "description": "Information regarding patient's living conditions.",
    "related_attributes": [
        {"name": "value", "description": "Place where the patient resides.", "examples": ["apartment", "old building"]},
        {"name": "status", "description": "The status is 'living' if the patient is currently living at the place or 'not_living' if the patient is not currently living at the place."},
        {"name": "household_size", "description": "Size of the patient's household.", "examples": ["2", "4"]}
    ]
}, {
    "slot": "occupation",
    "description": "Information regarding the patient's occupation.",
    "related_attributes": [
        {"name": "value", "description": "Job/occupation of the patient.", "examples": ["nurse", "student"]},
        {"name": "status", "description": "The status is 'true' if the patient works/worked at the above occupation or 'false' if the patient does/did not work at the above occupation."},
        {"name": "exposure", "description": "Are there any hazards/substances/dangers to which the patient got exposed at work?", "examples": ["chemical fumes", "dust"]}
    ]
}, {
    "slot": "travel",
    "description": "Information regarding the patient's recent travels.",
    "related_attributes": [
        {"name": "destination", "description": "Where has the patient travelled to?", "examples": ["canada", "united states"]},
        {"name": "status", "description": "The status is 'traveled' if the patient travelled recently or 'not_traveled' if the patient did not."},
        {"name": "date", "description": "When did the patient travel?", "examples": ["last week", "a year ago"]}
    ]
}, {
    "slot": "basic_information",
    "description": "Basic information about the patient.",
    "related_attributes": [
        {"name": "age", "description": "Age of the patient.", "examples": ["32", "50"]},
        {"name": "gender", "description": "Gender of the patient.", "examples": ["male", "female"]},
        {"name": "name", "description": "Name of the patient.", "examples": ["John", "Lily"]}
    ]
}]
```
IMPORTANT INSTRUCTIONS:
1. Read the given definitions carefully.
2. For a given dialogue history and last turn, only some of the intents, slots and related attributes are applicable.
3. Related attribute 'value' of the the slots symptom, medical_history, family_history, habit, exposure, medication and medical_test must be a standard medical concept.
4. Expected output should contain intent, slot and related values from the last turn. dialogue history is given as an additional context.

[dialogue history]
doctor: Uh itchy eyes, discharge anything like that?
patient: No, nothing like that.

[last turn]
doctor: About a stuffy or runny nose?
patient: No, nothing like that.

[output]
[{"intent": "inform", "slots": {"symptom": [{"value": "nasal congestion", "status": "negative"}, {"value": "rhinorrhea", "status": "negative"}]}}]
.
.
{{Remaining Exemplars}}
.
.
[dialogue history]
doctor: And um, have you had, uh, have you had any headaches?
patient: Uh, no headaches.

[last turn]
doctor: Any uh, stuffy nose or runny nose?
patient: Uh no, nothing like that.

[output]
\end{Verbatim}
\end{tcolorbox}

\subsection{POL Prompt}\label{app:pol_prompt}
\begin{tcolorbox}[
    enhanced, breakable, colback=white,
    colframe=black, boxrule=0.25mm,
    width=\textwidth
]
\fontsize{6pt}{6pt}\selectfont
You are a professional medical assistant who is an expert in understanding doctor-patient dialogues. The user will show you current state of the dialogue between a doctor and a patient and the last turn in their dialogue. Your task is to suggest the doctor's action as a continuation of the dialogue. Doctor's action consists of intents, slots and related attributes (if applicable). Definitions for intent, slots, and related attributes are given below as Python dictionaries.

\begin{Verbatim}[breaklines]
```
intents = [{
    "name": "inquire",
    "description": "The doctor is inquiring information from the doctor."
},
{
    "name": "chit-chat",
    "description": "The doctor is chit-chatting with the patient."
},
{
    "name": "nod_prompt_salutations",
    "description": "The doctor is nodding to the patient or delivering salutations."
},
{
    "name": "diagnosis",
    "description": "The doctor making a diagnosis."
},
{
    "name": "other",
    "description": "Any other action."
}]

slots = [{
    "slot": "symptom",
    "description": "A symptom relevant to the patient's condition.",
    "related_attributes": [
        {"name": "value", "description": "The symptom in medical terms.", "examples": ["coughing, dyspnea"]},
        {"name": "onset", "description": "When did this symptom appear?", "examples": ["three days ago", "one week back"]},
        {"name": "initiation", "description": "How did this symptom appear?", "examples": ["abruptly", "gradually"]},
        {"name": "location", "description": "Where is the symptom located?", "examples": ["back", "neck"]},
        {"name": "duration", "description": "How long does the symptom persist?", "examples": ["a few minutes", "a few hours"]},
        {"name": "severity", "description": "What is the severity of this symptom on a scale of 10?", "examples": ["4", "7"]},
        {"name": "progression", "description": "How is the symptom's progression?", "examples": ["getting worse", "constant"]},
        {"name": "frequency", "description": "Frequency, if applicable, to the symptom.", "examples": ["3-4 times a day", "every hour"]},
        {"name": "positive_characteristics", "description": "A characteristic positively associated with the symptom.", "examples": ["sharp", "burning"]},
        {"name": "negative_characteristics", "description": "A characteristic not associated with the symptom.", "examples": ["sharp", "burning"]},
        {"name": "unknown_characteristics", "description": "A characteristic with unknown relation with the symptom.", "examples": ["sharp", "burning"]},
        {"name": "alleviating_factor", "description": "A condition that alleviates the symptom.", "examples": ["laying down", "sleeping"]},
        {"name": "not_alleviating_factor", "description": "A condition that does not alleviate the symptom.", "examples": ["laying down", "sleeping"]},
        {"name": "aggravating_factor", "description": "A condition that aggravates the symptom.", "examples": ["laying down", "sleeping"]},
        {"name": "not_aggravating_factor", "description": "A condition that does not aggravate the symptom.", "examples": ["laying down", "sleeping"]},
        {"name": "not_alleviating_aggravating_factor", "description": "A condition that neither alleviates nor aggravates the symptom.", "examples": ["laying down", "sleeping"]},
        {"name": "unknown_factor", "description": "A condition with unknown alleviation/aggravation status.", "examples": ["laying down", "sleeping"]},
        {"name": "volume", "description": "Volume, if applicable to the symptom.", "examples": ["couple of teaspoons"]},
        {"name": "color", "description": "Color, if applicable to the symptom.", "examples": ["ping", "red"]},
        {"name": "itching", "description": "How severe is the itching on a scale of 10?", "examples": ["4", "7"]},
        {"name": "lesion_size", "description": "Is the lesion (or are the lesions) larger than 1cm (Yes/No)?"},
        {"name": "lesions_peel_off", "description": "Do the lesions peel off (Yes/No)?"},
        {"name": "rash_swollen", "description": "Is the rash swollen (Yes/No)?"}
    ]
}, {
    "slot": "medical_history",
    "description": "A medical condition relevant to the patient's medical history.",
    "related_attributes": [
        {"name": "value", "description": "Name of the medical condition.", "examples": ["hypertensive disease", "malignant neoplasm"]},
        {"name": "starting", "description": "When did the patient start to experience the condition?", "examples": ["since teenage", "ten years ago"]},
        {"name": "frequency", "description": "How frequently does the patient experience the added condition?", "examples": ["every year", "during summer"]}
    ]
}, {
    "slot": "family_history",
    "description": "A medical condition relevant to the patient's family.",
    "related_attributes": [
        {"name": "value", "description": "Name of the medical condition.", "examples": ["hypertensive disease", "malignant neoplasm"]},
        {"name": "relation", "description": "Relationship with the patient", "examples": ["mother", "aunt"]}
    ]
}, {
    "slot": "habit",
    "description": "An habitual activity such as smoking, alcoholism, etc.",
    "related_attributes": [
        {"name": "value", "description": "Name of an activity.", "examples": ["smoking", "marijuana"]},
        {"name": "starting", "description": "When did the patient pick up the activities?", "examples": ["ten years back", "as a child"]},
        {"name": "frequency", "description": "How frequently does the patient engage in the selected activity?", "examples": ["on weekends", "every day"]}
    ]
}, {
    "slot": "exposure",
    "description": "An environmental/chemical factor such as asbestos, pets, etc.",
    "related_attributes": [
        {"name": "value", "description": "Name of an environmental factor.", "examples": ["pets", "dust"]},
        {"name": "where", "description": "Where was the patient exposed to the selected factor?", "examples": ["work", "home"]},
        {"name": "when", "description": "When was the patient exposed to the selected factor?", "examples": ["four days ago"]}
    ]
}, {
    "slot": "medication",
    "description": "A medication.",
    "related_attributes": [
        {"name": "value", "description": "Name of a medication.", "examples": ["over-the-counter medicine", "paracetamol"]},
        {"name": "start", "description": "Since when did the patient start taking the medication?", "examples": ["few weeks ago", "two days back"]},
        {"name": "impact", "description": "Did the medication help the patient (Yes/No/Maybe)?"},
        {"name": "respone_to", "description": "For which condition/symptom is medication for?", "examples": ["hypertensive disease", "diabetes"]},
        {"name": "frequency", "description": "How frequently does the patient take the medication?", "examples": ["daily"]}
    ]
}, {
    "slot": "medical_test",
    "description": "A medical test.",
    "related_attributes": [
        {"name": "value", "description": "Name of a medical test.", "examples": ["chest X-ray", "electrocardiogram"]},
        {"name": "when", "description": "When did the patient had the medical test done?", "examples": ["yesterday", "a week ago"]}
    ]
}, {
    "slot": "residence",
    "description": "Information regarding patient's living conditions.",
    "related_attributes": [
        {"name": "value", "description": "Place where the patient resides.", "examples": ["apartment", "old building"]},
        {"name": "household_size", "description": "Size of the patient's household.", "examples": ["2", "4"]}
    ]
}, {
    "slot": "occupation",
    "description": "Information regarding the patient's occupation.",
    "related_attributes": [
        {"name": "value", "description": "Job/occupation of the patient.", "examples": ["nurse", "student"]},
        {"name": "exposure", "description": "Are there any hazards/substances/dangers to which the patient got exposed at work?", "examples": ["chemical fumes", "dust"]}
    ]
}, {
    "slot": "travel",
    "description": "Information regarding the patient's recent travels.",
    "related_attributes": [
        {"name": "destination", "description": "Where has the patient travelled to?", "examples": ["canada", "united states"]},
        {"name": "date", "description": "When did the patient travel?", "examples": ["last week", "a year ago"]}
    ]
}, {
    "slot": "basic_information",
    "description": "Basic information about the patient.",
    "related_attributes": [
        {"name": "age", "description": "Age of the patient.", "examples": ["32", "50"]},
        {"name": "gender", "description": "Gender of the patient.", "examples": ["male", "female"]},
        {"name": "name", "description": "Name of the patient.", "examples": ["John", "Lily"]}
    ]
}, {
    "slot": "disease",
    "description": "A medical condition.",
    "related_attributes": [
        {"name": "value", "description": "Name of the medical condition", "examples": ["viral pneumonia", "common cold"]}
    ]
}]
```
IMPORTANT INSTRUCTIONS:
1. Read the given definitions carefully.
2. For a given dialogue state and last turn, only some of the intents, slots and related attributes are applicable.
3. Related attribute 'value' of the the slots symptom, medical_history, family_history, habit, exposure, medication and medical_test must be a standard medical concept.
4. Make sure that the doctor's action is a continuation of the dialogue. dialogue state is given as an additional context.

[dialogue state]
```
{
  "positive_symptom": [
    {
      "value": "pharyngitis",
      "onset": "past four days"
    },
    {
      "value": "fever",
      "onset": "last two days",
      "positive_characteristics": [
        "during the day"
      ]
    },
    {
      "value": "deglutition disorders"
    },
    {
      "value": "erythema",
      "location": [
        "pharyngeal structure"
      ]
    },
    {
      "value": "body substance discharge",
      "location": [
        "pharyngeal structure"
      ],
      "color": "whitish"
    },
    {
      "value": "swelling",
      "location": [
        "left lateral part of neck",
        "neck",
        "right lateral part of neck"
      ]
    },
    {
      "value": "lymphadenopathy",
      "location": [
        "left lateral part of neck",
        "neck",
        "right lateral part of neck"
      ]
    },
    {
      "value": "chills"
    }
  ],
  "negative_symptom": [
    {
      "value": "dysphonia"
    },
    {
      "value": "night sweats"
    },
    {
      "value": "headache"
    }
  ],
  "avail_medical_test": [
    {
      "value": "body temperature measurement",
      "when": "last night"
    }
  ]
}
```

[last turn]
doctor: OK. Uhm, and have you had any headaches?
patient: No headaches.

[output]
[{"action": "inquire", "symptom": [{"value": "redness of eye"}]}, {"action": "inquire", "symptom": [{"value": "body substance discharge", "checks": [{"type": "location", "values": ["eye"]}]}]}]
.
.
{{Remaining Exemplars}}
.
.
[dialogue state]
```
{
  "positive_symptom": [
    {
      "value": "pharyngitis",
      "onset": "last week"
    },
    {
      "value": "chills",
      "onset": "last few nights"
    },
    {
      "value": "bedridden",
      "onset": "a week ago"
    },
    {
      "value": "deglutition disorders"
    }
  ],
  "negative_symptom": [
    {
      "value": "headache"
    }
  ]
}
```

[last turn]
doctor: And um, have you had, uh, have you had any headaches?
patient: Uh, no headaches.

[output]
\end{Verbatim}
\end{tcolorbox}

\subsection{NLG Prompt}\label{app:nlg_prompt}
\begin{tcolorbox}[
    enhanced, breakable, colback=white,
    colframe=black, boxrule=0.25mm,
    width=\textwidth
]
\fontsize{6pt}{6pt}\selectfont
You are a professional medical assistant who is an expert in understanding doctor-patient dialogues. The user will show you the last turn of the dialogue between a doctor and a patient and the doctor's action. Your task is to suggest the doctor's response as a continuation of the dialogue.

\begin{Verbatim}[breaklines]
IMPORTANT INSTRUCTIONS:
1. Your suggested response must reflect the doctor's actions and form a natural continuation of the dialogue.
2. Your suggested response must be fluent, grammatically correct and empathetic.
3. Your suggested response must satisfy any queries made by the user.

[actions]
```
[
  {
    "action": "inquire",
    "symptom": [
      {
        "value": "redness of eye"
      }
    ]
  },
  {
    "action": "inquire",
    "symptom": [
      {
        "value": "body substance discharge",
        "checks": [
          {
            "type": "location",
            "values": [
              "eye"
            ]
          }
        ]
      }
    ]
  }
]
```

[last turn]
doctor: OK. Uhm, and have you had any headaches?
patient: No headaches.

[output]
Have you had any eye redness or eye discharge?
.
.
{{Remaining Exemplars}}
.
.
[actions]
```
[
  {
    "action": "inquire",
    "symptom": [
      {
        "value": "nasal congestion"
      },
      {
        "value": "rhinorrhea"
      }
    ]
  }
]
```

[last turn]
doctor: And um, have you had, uh, have you had any headaches?
patient: Uh, no headaches.

[output]
\end{Verbatim}
\end{tcolorbox}
\twocolumn

\end{document}